\documentclass[lettersize,journal]{IEEEtran}
\usepackage{cite}
\usepackage{amsmath,amssymb,amsfonts}
\usepackage{algorithmic}
\usepackage{graphicx}
\usepackage{textcomp}
\usepackage{booktabs}
\usepackage{amssymb}
\usepackage{epstopdf} 
\usepackage{bbding} 
\usepackage{tabularx,threeparttable}
\usepackage{array}
\usepackage[colorlinks,linkcolor=blue]{hyperref}
\usepackage{mathtools}
\usepackage{mathrsfs}
\usepackage{multirow}
\usepackage{url}
\usepackage{color}
\usepackage{dsfont}
\usepackage{subcaption}
\usepackage{caption}
\usepackage{subcaption}
\usepackage{colortbl} 
\definecolor{lightgray}{gray}{0.85} 
\newif\ifshowchanges
\showchangestrue

\begin{document}
\title{Mamba-Sea: A Mamba-based Framework with Global-to-Local Sequence Augmentation for Generalizable Medical Image Segmentation}
\author{Zihan Cheng, Jintao Guo, Jian Zhang, Lei Qi, Luping Zhou, Yinghuan Shi$^*$, Yang Gao
\thanks{This work is supported by NSFC Project (62222604, 62206052), China Postdoctoral Science Foundation (2024M750424), Fundamental Research Funds for the Central Universities (020214380120, 020214380128), State Key Laboratory Fund (ZZKT2024A14), Postdoctoral Fellowship Program of CPSF (GZC20240252), Jiangsu Funding Program for Excellent Postdoctoral Talent (2024ZB242) and Jiangsu Science and Technology Major Project (BG2024031).}
\thanks{Zihan Cheng is with the Shanghai Jiao Tong University School of Medicine, Shanghai 200025, China; and also with the National Institute of Healthcare Data Science, Nanjing University, Nanjing, Jiangsu 210093, China. (e-mail: chengzihan@sjtu.edu.cn). }
\thanks{Jintao Guo, Jian Zhang, Yinghuan Shi and Yang Gao are with the State Key Laboratory for Novel Software Technology and the National Institute of Healthcare
Data Science, Nanjing University, Nanjing, Jiangsu 210093, China. (e-mail: guojintao@smail.nju.edu.cn; zhangjian7369@smail.nju.edu.cn; syh@nju.edu.cn; gaoy@nju.edu.cn).}
\thanks{Lei Qi is with the School of Computer Science and Engineering, Key Lab of Computer Network and Information Integration, Southeast University, Nanjing, Jiangsu 211189, China. (e-mail: qilei@seu.edu.cn).}
\thanks{Luping Zhou is with the School of Electrical and Information
Engineering, The University of Sydney, Sydney, NSW 2006, Australia. (e-mail: luping.zhou@sydney.edu.au).}
\thanks{The corresponding author of this work is Yinghuan Shi.}}
\maketitle

\begin{abstract}
To segment medical images with distribution shifts, domain generalization (DG) has emerged as a promising setting to train models on source domains that can generalize to unseen target domains. 
Existing DG methods are mainly based on CNN or ViT architectures. 
\textcolor{black}{Recently, advanced state space models, represented by Mamba, have shown promising results in various supervised medical image segmentation.} 
The success of Mamba is primarily owing to its ability to capture long-range dependencies while keeping linear complexity with input sequence length, making it a promising alternative to CNNs and ViTs. 
Inspired by the success, in the paper, we explore the potential of the Mamba architecture to address distribution shifts in DG for medical image segmentation.
Specifically, we propose a novel Mamba-based framework, \textbf{Mamba-Sea}, incorporating global-to-local sequence augmentation to improve the model's generalizability under domain shift issues. 
\textcolor{black}{Our Mamba-Sea introduces a global augmentation mechanism designed to simulate potential variations in appearance across different sites, aiming to suppress the model's learning of domain-specific information.}
\textcolor{black}{At the local level, we propose a sequence-wise augmentation along input sequences, which perturbs the style of tokens within random continuous sub-sequences by modeling and resampling style statistics associated with domain shifts.} 
To our best knowledge, Mamba-Sea is the first work to explore the generalization of Mamba for medical image segmentation, providing an advanced and promising Mamba-based architecture with strong robustness to domain shifts. 
Remarkably, our proposed method is the first to surpass a Dice coefficient of 90\% on the Prostate dataset, which exceeds previous SOTA of 88.61\%. The code is available at \href{https://github.com/orange-czh/Mamba-Sea}{https://github.com/orange-czh/Mamba-Sea}.
\end{abstract}

\begin{IEEEkeywords}
Medical Image Segmentation; Domain Generalization; State Space Model; Mamba.
\end{IEEEkeywords}

\section{Introduction}
\label{sec:introduction}
\IEEEPARstart{R}{ecently}, deep learning methods have achieved remarkable performance in medical image segmentation, and most existing models are built on the assumption that the training and testing data are identically and independently distributed, \textit{i.e.},  \textbf{i.i.d}\cite{ronneberger2015u,1,goudarzi2023segmentation}. 
However, in clinical practice, the factors \textit{e.g.} variations in scanners, imaging protocols, and operators can lead to significant distribution differences between training and testing data, which can severely impact model performance and hinder the generalization of segmentation models across different clinical environments\cite{liu2020shape,gu2023cddsa}.
\begin{figure}[!t]
\centerline{\includegraphics[width=\columnwidth]{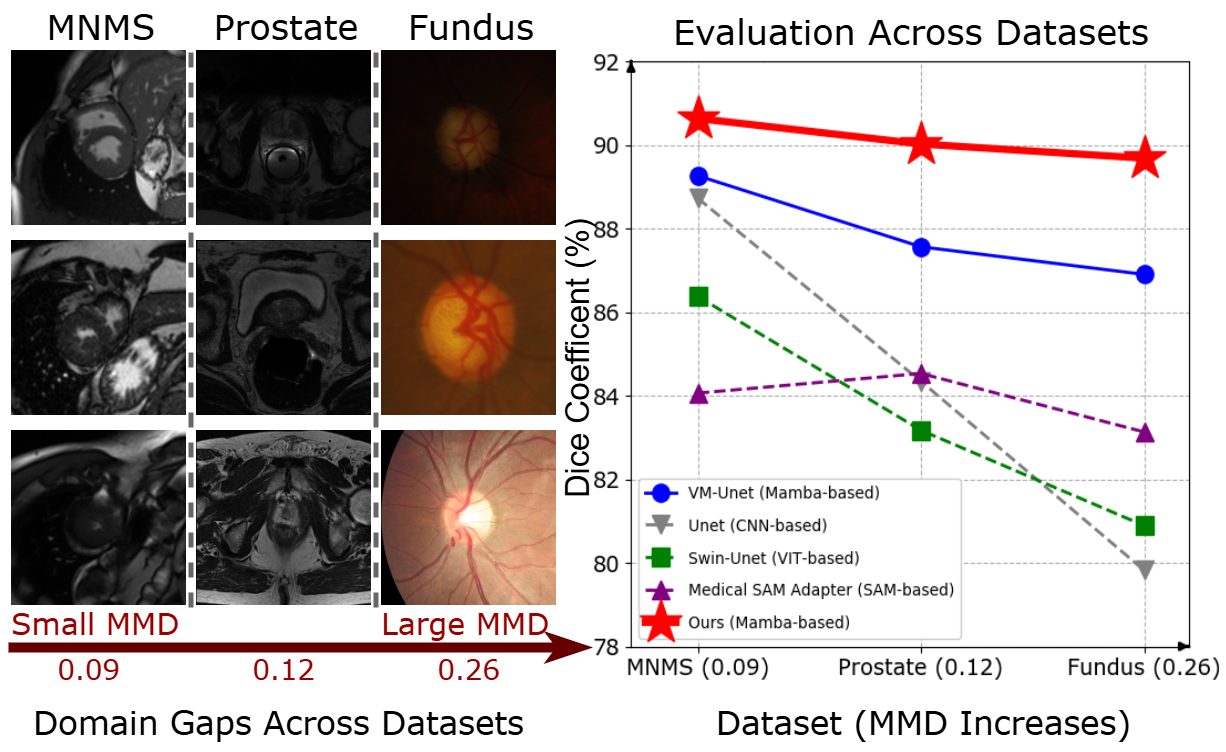}}
\caption{Typical images from three public medical datasets with
increasing domain gaps; and comparison of the segmentation performance of Mamba-Sea with UNet\cite{ronneberger2015u}, Swin-Unet\cite{2}, \textcolor{black}{Med-SA\cite{wu2023medical}}, and VM-Unet\cite{ruan2024vm} on these three datasets.}
\label{compare}
\end{figure}

To segment medical images with distribution shifts, domain generalization (DG) has been explored, where models are trained on source domains and aim to generalize to unseen target domains without retraining. 
\textcolor{black}{Existing DG methods in medical image segmentation mainly include meta-learning\cite{liu2020shape,liu2021semi}, data augmentation or manipulation\cite{8,9}, and feature alignment or disentanglement\cite{13,15,16}. 
Despite these strategies helping the model learn robust representations that enhance performance across domains, they are mainly based on CNN architectures, which have limited receptive fields due to local convolutions. Consequently, the CNN-based methods tend to learn local texture information, leading to overfitting to source domains\cite{bai2022improving}. Recent works in DG for medical image segmantation have introduced Vision Transformers (ViTs) as the backbone, utilizing the global receptive field of the self-attention mechanism to mitigate local texture bias\cite{kim2023dimix}. However, the complexity of self-attention increases quadratically with input length, leading to considerable computational overhead for ViTs when modeling long sequences\cite{shamshad2023transformers}.}

To address the issue, modern state-space models (SSMs), represented by Mamba\cite{gu2023mamba}, has been emerged and proved the ability of SSMs to model long-range dependencies among tokens while maintaining linear complexity relative to input size\cite{dang2024log, hatamizadeh2024mambavision, shi2024vssd,yue2024medmamba,nasiri2024vision}.
Some pioneers have proposed Mamba-based methods in enhancing performance for medical image segmentation\cite{wang2024mamba,xu2024polyp,xing2024segmamba,xu2024hc,wu2024ultralight}, which these studies are predominantly focused on fully supervised tasks in medical image segmentation, typically relying on the \textit{i.i.d} assumption that the source (training) and target (testing) domains follow the same distribution. 
Few works investigate the generalization ability of Mamba on unseen domains that diverge from the source domain distribution. 
Thus it remains an open question whether Mamba could achieve promising performances in the challenging DG tasks of medical image segmentation. 
The core challenge lies in how to effectively constrain the model to learn domain-invariant features to fight against domain shift during the state space modeling process.

To solve the problem, we first investigate the generalization of Mamba by comparing its performance with CNN- and ViT-based models.
Specifically, we analyze the segmentation performance of the advanced Mamba model, \textit{i.e.}, VM-Unet\cite{ruan2024vm}, comparing it with the representative models in medical image segmentation, including the CNN-based Unet\cite{ronneberger2015u} and ViT-based Swin-Unet\cite{2}. 
We also perform a comparison with the representative foundation model, Medical SAM Adaptation (Med-SA) \cite{wu2023medical}, which finetunes SAM\cite{kirillov2023segment} using LoRA\cite{hu2021lora} and relies on a large number of parameters and substantial additional training data to achieve robust performance in broad medical image segmentation tasks. 
The experiments are conducted on three representative medical image segmentation datasets, where domain gap increases across the datasets (MNMS \cite{campello2021multi}, Prostate \cite{liu2020shape}, Fundus \cite{14}). 
Following \cite{tolstikhin2016minimax}, \textcolor{black}{domain gap is quantified} by calculating mean of Maximum Mean Discrepancy (MMD) values between each pair of domains. A small MMD value indicates a small domain gap.

As illustrated in Fig. \ref{compare}, we observe that (1) \textit{VM-Unet\cite{ruan2024vm} consistently demonstrates improvements across all datasets in comparison to CNN- and ViT-based models}. 
Furthermore, \textit{VM-Unet significantly surpasses Med-SA on three representative medical image segmentation datasets with domain shifts}.
The effectiveness can be attributed to the selective scan mechanism of VMamba\cite{liu2024vmamba}, which captures global semantic information, and the U-shape architecture combined with skip connections, which excels in capturing local fine-grained pixel-level dependencies\cite{ma2024u}.
(2) \textit{The performance of existing methods based on CNNs and ViTs exhibits a marked decline as domain gap expands, while Mamba demonstrates a relatively smaller decrease in performance.} This finding underscores the potential of the SSM-based architecture of Mamba for effectively learning domain-invariant information for medical image segmentation. 
(3) Nevertheless, \textit{as the domain gap increases, Mamba’s performance still declines to some extent.}
It can be attributed to the input-dependent matrixes of the Mamba model inevitably accumulate domain-specific information during training\cite{guo2024start}.
In light of these observations, we aim to investigate an effective Mamba-based framework that sufficiently incentivizes the generalization of Mamba for DG medical image segmentation tasks.

In our work, we introduce a generalizable \textcolor{black}{\textbf{Mamba}}-based framework with global-to-local \textcolor{black}{\textbf{Se}}quence \textcolor{black}{\textbf{A}}ugmentation, namely \textbf{Mamba-Sea}, which leverages Mamba's strengths in sequential modeling. 
The global-to-local sequence augmentation method aims to explicitly constrain the model to learn domain-invariant features by focusing on Mamba's capability to model dependencies among tokens sequentially while maintaining a global receptive field.
This method introduces diverse perturbations into both the \textit{global information spanning the entire image} and the \textit{local dependencies among different sequences}.
At the global level, we introduce a global appearance variation augmentation method, abbreviated as \textbf{GVA}, which simulates potential variations across different sites. This method utilizes a learnable lightweight network to generate diverse global perturbations for the entire image.
Meanwhile, at the local level, we find that Mamba relies on multi-directional scanning to establish pixel-level dependencies, implicitly encouraging the model to learn domain-specific information embedded in local regions. 
To address this issue, we employ a local sequence-wise style transformation augmentation abbreviated as \textbf{LSA}, to improve the learning of semantic information within localized regions. 
LSA aims to simulate domain shifts during training by introducing style perturbations to sequences focused on by the input-dependent matrices and model the uncertainty of feature statistics associated with domain shifts. By combining LSA and GVA, our framework incorporates global-to-local sequence augmentation to enhance the model’s generalizability under domain shifts in medical image segmentation.

Our contribution can be summarized as follows:
\begin{itemize}
\item We propose Mamba-Sea, a novel Mamba-based generalizable framework for DG medical image segmentation, which exhibits strong performance and serves as a competitive alternative to CNNs and ViTs. 
To the best of our knowledge, Mamba-Sea is the first work to study the cross-domain generalizability of Mamba in medicine.  
\item We design an effective global-to-local sequence augmentation that simulates diverse potential domain shifts in modeling sequence token dependencies. This method can enhance the model learning of global semantic features and local pixel-level dependency, thus effectively fully stimulating the generalizability of Mamba architecture.
\item We explore a semantic consistency training strategy to enforce prediction consistency across different perturbations, thereby improving the model's ability to learn domain-invariant features and enhancing its robustness to domain shifts.
\end{itemize}

\begin{figure*}[!t]
\centerline{\includegraphics[width=\textwidth]{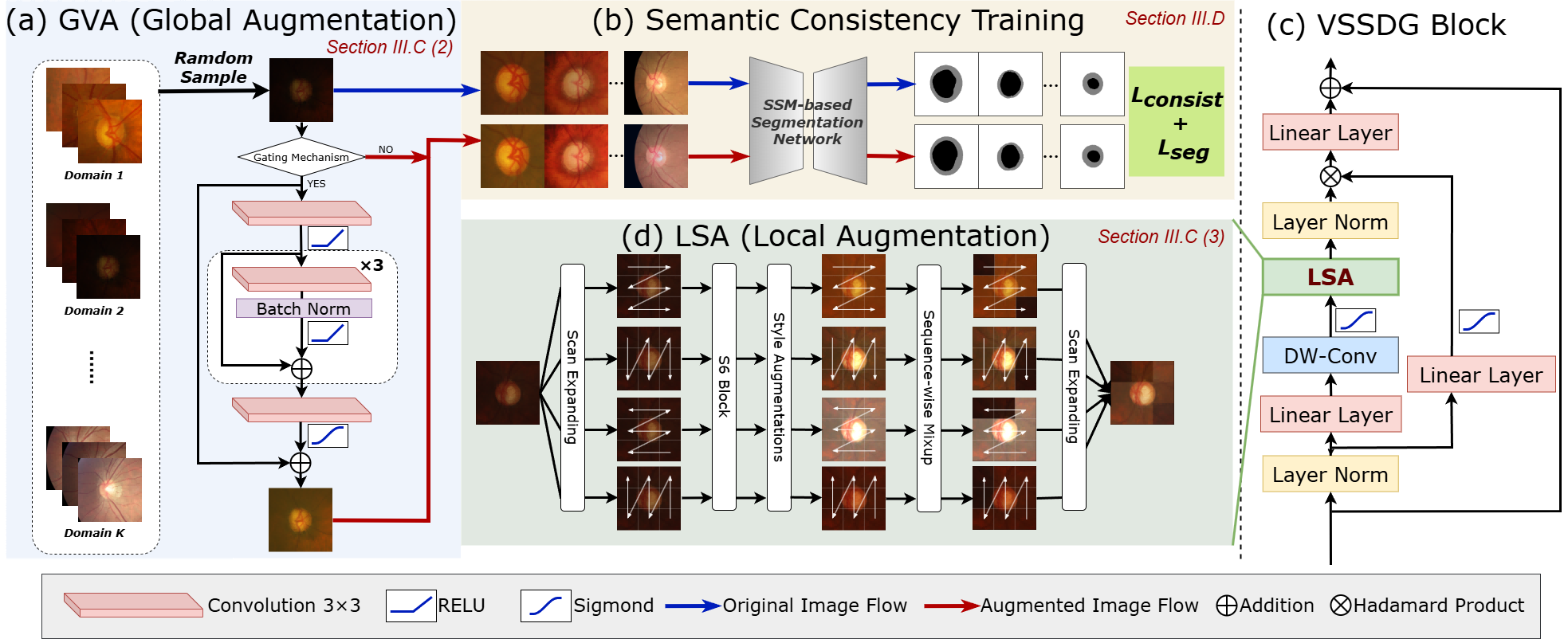}}
\caption{The overall structure of Mamba-Sea. (a) represents \textcolor{black}{GVA} module. As shown in (b), both the original images and the globally-augmented images are fed into the SSM-based generalizable segmentation network for training. (c) represents the core components of segmentation network, where we design \textcolor{black}{LSA} module shown in (d) to achieve augmentation at the local level.}
\label{fig1}
\end{figure*}

We conduct expensive experiments on two publicly available DG medical image segmentation datasets \textcolor{black}{and a large-scale dataset for skin lesion segmentation}. The results demonstrate that our method achieves state-of-the-art (SOTA) performance, \textit{e.g.}, our Mamba-Sea surpasses the best CNN-based method\cite{hu2022domain} by $1.41\%$ ($90.02\%$ \textit{vs.} $88.61\%$) on the Prostate dataset and surpasses the best ViT-based method\cite{kim2023dimix} by $1.14\%$ ($89.68\%$ \textit{vs.} $88.54\%$) on the Fundus dataset.

\section{Related Works}
\subsection{Domain Generalized Medical Image Segmentation}
In recent years, domain generalization (DG) has garnered significant attention in recent years aimed at addressing segmentation under domain shift, particularly in the field of medical image segmentation. 
DG-based segmentation methods for medical images can be roughly classified into three categories. 
The first category is \textbf{meta-learning} methods, which aim to enhance generalization by dividing source domains into meta-train and meta-test subsets and using meta-optimization to iteratively adapt the model's parameters across the meta-train tasks. 
For example, Liu \textit{et al.} combine meta-learning with federated learning to enable privacy-preserving and generalizable segmentation\cite{liu2021feddg}. 
Dou \textit{et al.} propose MASF which aligns class relationships and clusters class-specific features for cross-site brain segmentation\cite{dou2019domain}. 
The second category includes methods focused on \textbf{data} augmentation or manipulation techniques. They increase diversity and quantity of the original training data, ultimately improving the model’s adaptability to different domains. 
\textcolor{black}{Typically, Li \textit{et al.} augment a domain by modifying cardiac images with sampled shape and spatial\cite{li2021random}. Tomar \textit{et al.} combine knowledge distillation with adversarial-based data augmentation for cross-site medical image segmentation\cite{tomar2023tesla}.} 
The third category extracts domain-invariant \textbf{features} from input images. These methods aim to learn a feature representation that remains consistent across multiple domains. 
For instance, Wang \textit{et al.} incorporate a domain knowledge pool to learn domain prior information extracted from multi-source domains\cite{14}. Hu \textit{et al.} propose DCAC where a dynamic convolutional head is conditioned on global image features to adapt to unseen target domains\cite{hu2022domain}.

The proposed methods have demonstrated superior performance. However, despite their success, we observe that these approaches primarily utilize segmentation frameworks based on CNNs or ViTs. 
To the best of our knowledge, Mamba-Sea is the first work  to employ Mamba for DG tasks in medical image segmentation. 
By integrating both global and local strategies, we explore the potential of Mamba in domain-generalizable medical image segmentation.

\subsection{Mamba for Medical Image Segmentation}
Modern state-space models, represented by Mamba, have demonstrated promising performance in medical image segmentation for their ability to effectively capture long-range dependencies while maintaining linear complexity\cite{gu2023mamba}. 
Recent studies have validated Mamba’s efficacy in medical image segmentation tasks. 
We enumerate several representative cases. For example, Wang \textit{et al.} introduce Mamba-UNet, which esynergizes the U-Net in medical image segmentation with Mamba's capability\cite{wang2024mamba}. 
\textcolor{black}{Xing \textit{et al.} introduce SegMamba, a 3D medical image segmentation model that captures long-range dependencies in processing speed and whole volume feature modeling\cite{xing2024segmamba}.}
Zhang \textit{et al.} propose VM-UNetV2 which has the ability to capture extensive contextual information and integrates low- and high-level features in medical images\cite{zhang2024vm}. 
\textcolor{black}{Wang \textit{et al.} design LKM-UNet which partitions images into pixel-level segments and applys bidirectional Mamba to achieve large kernel spatial modeling\cite{wang2024large}.}
\textcolor{black}{Archit \textit{et al.} develop LKM-UNet, which achieves global feature extraction without the need for the skip connections\cite{archit2024vim}.}
\textcolor{black}{Khan \textit{et al.} propose CAMS-Net, a convolution and attention-free segmentation Mamba-based network for medical images\cite{khan2024convolution}.}

Nevertheless, these methods are applied primarily to traditional medical segmentation tasks without considering the domain shifts. Medical image datasets often contain source domains that do not fully encompass all possible variations.  
As aforementioned, Mamba has shown potential in combating distribution shifts. 
Taking into account the structural characteristic of selective scanning mechanism across sequences in Mamba, we propose a global-to-local sequence augmentation module designed to fully leverage Mamba's generalizability against domain shifts in medical image segmentation.

\section{Method}
\subsection{Definition and Overview}
The domain generalization task is defined as follows: given $K$ observed source domains $\mathcal{D}_S = \{D_s^1, D_s^2, ..., D_s^K\}$ that follow different distributions.
For each domain, $D_s^k = \{(x_p^k, y_p^k)\}_{i=1}^{n_k}$, where $n_k$ is the number of samples in $D_s^k$, and $(x_p^k, y_p^k)$ denotes the sample-label pair for the $p$-th sample in the $k$-th domain. 
The goal of domain generalization is to use the source domains $\mathcal{D}_S$ to train a model that performs well on unseen target domains $\mathcal{D}_T$ with different distributions.

As shown in Fig. \ref{fig1}, we propose Mamba-Sea, a Mamba-based framework for generalizable medical image segmentation that incorporates global-to-local augmentation at sequences.  
Our Mamba-Sea framework consists of three main stages. 
1) \textcolor{black}{Firstly, the input images are processed through the GVA module,} which simulates variations in the appearance of medical images from different sites to apply global augmentation.
2) Secondly, both the globally augmented images and the original images are fed into our SSM-based generalizable segmentation network \textcolor{black}{based on VM-UNet\cite{ruan2024vm} for joint training. 
This network incorporates the LSA module}, which diversifies training data by reparameterizing and mixing features based on their statistics and introduces variability by applying a probabilistic mask to blend the transformed and original features. 
3) \textcolor{black}{Finally, a semantic training strategy is designed to promote prediction consistency between the outputs of source domain images and augmented images. A consistency loss is used to encourage the model to produce similar segmentation outputs for both types of images,} further strengthening its resistance to domain shifts. 
We discuss all components comprehensively in the following.

\subsection{Preliminaries}
\label{sub:preliminaries}
Contemporary SSM-based models such as Structured State Space Sequence Models\cite{gu2021efficiently} and Mamba\cite{gu2023mamba}, are inspired by the traditional continuous system. 
We define $f_h(\cdot)$ to denote the features extracted from the middle layer of SSM-based models.
After scan expansion, the input sequence $f_h(x)$ is reshaped from $\mathbb{R}^{B \times C \times W \times H} $ to $\mathbb{R}^{B \times D \times L}$, where $x$ denotes the batch of images fed into the model, $D$ represents the feature dimension and $L$ denotes the sequence length. 
This system processes a 1-D input sequence \( f_h(x_t) \in \mathbb{R}^{B \times D} \) through implicit intermediate states \( h_t \in \mathbb{R}^{B \times N} \) to produce \( f'_h(x_t) \in \mathbb{R}^{B \times D} \), where $t$ means time step. This transformation is governed by a linear Ordinary Differential Equation:

\begin{equation}
\begin{aligned}
    h'_t = Ah_t + Bf_h(x_t), f'_h(x_t) = Ch_t,
    \end{aligned}
\label{eq1}
\end{equation}
here, $A \in \mathbb{R} ^ {N \times N}$ represents the state matrix, while the projection parameters \(B \in \mathbb{R}^{N \times D}\) and \(C\in \mathbb{R}^{D \times N}\) respectively facilitate the mapping of the input to the hidden state and the hidden state to the output.

S4 and Mamba discretize the continuous system to align with deep learning applications. They introduce a timescale parameter $\Delta$ and convert the continuous parameters $A$ and $B$ into discrete parameters $\overline{A}$ and $\overline{B}$ using a fixed discretization rule. Typically, the zero-order hold method is used for this discretization, defined as follows:

\begin{equation}
\begin{aligned}
    \overline{A} = e^{\Delta A}, \overline{B} = (\Delta A)^{-1}(e^{\Delta A}- I)\cdot \Delta B.
\label{eq2}
\end{aligned}
\end{equation}

After the discretization of $\overline{A}$, $\overline{B}$, the discretized version of Eq. \eqref{eq1} using a step size $\Delta$ can be rewritten as:

\begin{equation}
\begin{aligned}
    h'_t = \overline{A}h_{t-1} + \overline{B}f_h(x_t), f'_h(x_t) = Ch_t.
\end{aligned}
\label{eq:linear_recurrence}
\end{equation}

Lastly, the output $f'_h(x) \in \mathbb{R}^{B \times D \times L}$ is computed through a global convolution:
\begin{equation}
\begin{aligned}
    \overline{K} = (C\overline{B}, C\overline{AB}, \ldots, C\overline{A}^{L-1}\overline{B}), f'_h(x) = f_h(x) * \overline{K},
\end{aligned}
\label{eq:global_convolution}
\end{equation}where $\overline{K} \in \mathbb{R}^{D \times L}$ represents a structured convolutional kernel, $L$ denotes the length of the input sequence $f_h(x)$, and $*$ denotes the convolution operation.

\subsection{Global-to-local Sequence Augmentations}
\subsubsection{Overview}
Considering the structural characteristics of selective scanning mechanism across sequences in Mamba, \textcolor{black}{we design global-to-local sequence augmentations which comprise a global-level module named GVA and a local-level module known as LSA}. Specially, \textbf{at the global level}, we employ a global augmentation module by employing a gating mechanism and a lightweight network. This approach adaptively simulates potential appearance variations by adjusting parameters across different convolutional layers.
GVA generates diverse global perturbations that span the entire image, effectively mimicking a wide range of domain shifts. 
\textcolor{black}{It provides diverse types of domain variations for the subsequent sequential scanning mechanism of Mamba, thereby enhancing our model's ability to generalize effectively in a global context.}
\textbf{At the local level}, 
\textcolor{black}{considering the multi-directional sequence scanning mechanism of SS2D in Mamba, we employ augmentations that synthesize feature statistics across varying sequences.} This strategy models the uncertainty associated with domain shifts to help boost our framework's generalizability across different domains.
By leveraging a global-local collaborative approach, the model's overfitting to source domains is alleviated, strengthening its capacity to learn domain-invariant information.

\begin{figure}[!t]
\centerline{\includegraphics[width=0.5\textwidth]{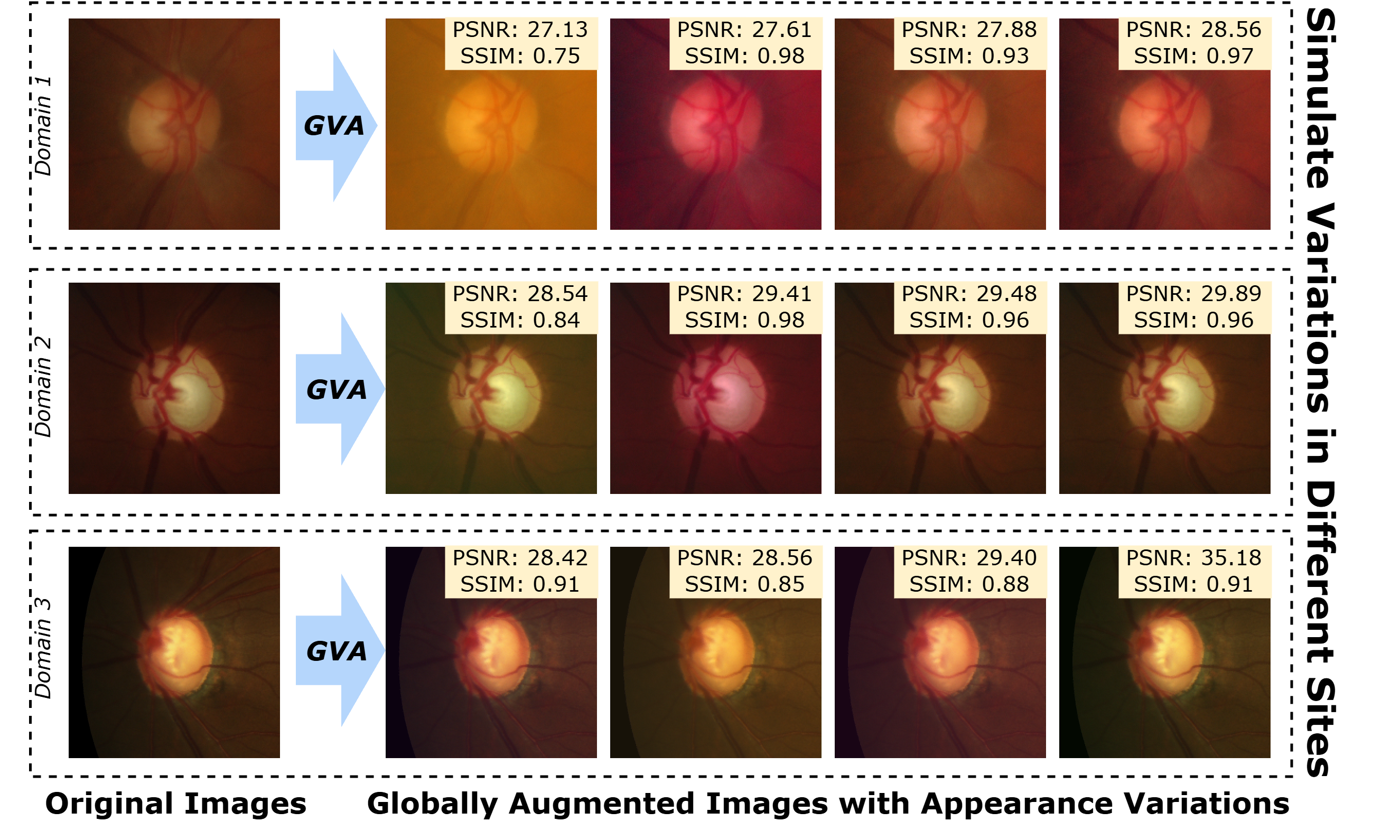}}
\caption{Examples of original images alongside their augmented images after applying our proposed global appearance variation augmentation (GVA) on \textbf{Fundus} dataset. The PSNR and SSIM values indicate the pixel-wise and structural similarities between the original and augmented images.
}
\label{globalvis}
\end{figure}

\subsubsection{GVA: Global Appearance Variation Augmentation}
In the field of medical imaging, images from different healthcare institutions often exhibit significant variations in appearance\cite{14}. 
Such inconsistencies pose challenges for medical image segmentation tasks, as segmentation models are typically trained under the assumption of \textit{i.i.d}\cite{isensee2021nnu}. 
Specially, we design GVA to simulate potential variations in different sites. 
GVA utilizes adaptive enhancement techniques to simulate different appearance by adjusting the overall illumination of images at the global level. 
\textcolor{black}{Fig. \ref{globalvis} is shown as an illustrative example. GVA is trained alongside the segmentation model. For the input images, GVA does not alter the dimensionality of the input. As a result, we convert the GVA outputs during training process into images to visually demonstrate the diversity generated by the globally-augmented process.
PSNR quantifies the pixel-level fidelity between the generated and original images, with higher values indicating closer resemblance. SSIM evaluates perceptual similarity by assessing structural information, luminance, and contrast, where higher scores represent better structural alignment. These metrics are used to demonstrate the diversity and quality of our globally-augmented images\cite{gu2020medsrgan,gourdeau2022proper}.}
Instead of merely focusing on brightness normalization, it dynamically augments the input image to create diverse visual conditions across domains. This process helps maintain consistent visual characteristics and enhances Mamba's ability to learn domain-invariant information during subsequent sequential scanning across different pathways, while exposing the model to multiple simulated conditions.

Specifically, inspired by low-light enhancement techniques in low-level vision\cite{ma2022toward}, we propose GVA which consists of a gating mechanism and a learnable lightweight network. 
The gating mechanism automatically determines if further augmentation is required. 
The lightweight network includes an input convolutional layer, multiple enhancement blocks, and an output convolutional layer, all of which perform global image enhancement using adaptive techniques.
The architecture is designed to be lightweight, allowing it to be easily integrated into existing medical image segmentation pipelines without an extra computational burden. The architecture of GVA is illustrated in Fig. \ref{fig1}(a).
GVA incorporates an adaptive augmentation strategy based on the overall brightness of the input image. 
In our GVA and LSA module, we do not distinguish between different domains. 
Therefore, we simplify the notation by omitting the superscript of \(x_p^k\), referring to the image simply as \(x_i\), $i \in \{1, 2, 3, …,B\}$.
Here, \(x_i\) represents the \(i\)-th image in the batch \(x\), which consists of \(B\) images. 
We define $f_g(\cdot)$ to denote the features extracted from the images using GVA. As a result, \(f_g(x) \in \mathbb{R}^{B \times C \times H \times W}\) and \(f_g(x_i) \in \mathbb{R}^{C \times H \times W}\). 
During the forward pass, we compute the mean brightness of \(x_i\), denoted as \(I(x_i)\), using the following equation:

\begin{equation}
\begin{aligned}
    I(x_i) = \frac{1}{C \cdot H \cdot W} \sum_{c=1}^{C} \sum_{h=1}^{H} \sum_{w=1}^{W} f_g(x_i^{c,h,w}),
\end{aligned}
\label{light}
\end{equation} 
here, $C$ is the number of channels, $H$ and $W$ denote the dimensions of height and weight, respectively.
Then we introduced $\mathcal{M} \in \mathbb{R}^{B \times 1}$ as a mask to selectively apply the augmented network, denoted as $\Phi(\cdot)$. $\mathcal{M}$ is defined as follows:
\begin{equation}
\begin{aligned}
\mathcal{M}_i =  \mathbb{I}[I(x_i) < \tau],
\end{aligned}
\label{mask_global}
\end{equation}
here \( \mathbb{I}(\cdot) \) is the indicator function that equals 1 if the condition is satisfied and 0 otherwise, $\tau$ represents the threshold for applying the globally augmented network $\Phi(\cdot)$. We expand $\mathcal{M}$ to $\mathbb{R}^{B \times C \times H \times W}$ to accommodate the dimensions of images. The input $f_g(x)$ is transformed to the globally-augmented $f'_g(x)$ as follows:
\begin{equation}
\begin{aligned}
f'_g(x) = \Phi(f_g(x)) \odot \mathcal{M} + f_g(x) \odot (1 - \mathcal{M}),
\end{aligned}
\label{globalmask}
\end{equation} 
where \( \odot \) denotes element-wise multiplication. $f'_g(x) \in \mathbb{R}^{B \times C \times H \times W}$ represents the feature of $B$ globally-augmented images. Specially, GVA transforms the original image $x_p^k$ to the globally-augmented image $x_{p,E}^k$.
As we analyse that well-illuminated images have sufficient edge information, so applying global augmentation may weaken their edge information. 
By preventing unnecessary processing of well-illuminated images, the mechanism enhances the network's efficiency.

GVA performs global augmentation by modifying the entire image's conditions to simulate variations of medicial images from different sites.
\textcolor{black}{To simplify the complexity of training, we train the GVA in conjunction with the subsequent segmentation network, both of which share the same loss function for supervisory signals.}
This module enables the network to simulate a broad range of potential conditions encountered in real-world medical imaging by efficiently expanding through parameter adjustments. 
By exposing segmentation network to various simulated scenarios, GVA provides diverse types of domain variations for the subsequent sequential scanning mechanism of Mamba, \textcolor{black}{thereby enhancing the model’s ability to generalize effectively in a global context}. 

\begin{figure}[!t]
\centerline{\includegraphics[width=0.5\textwidth]{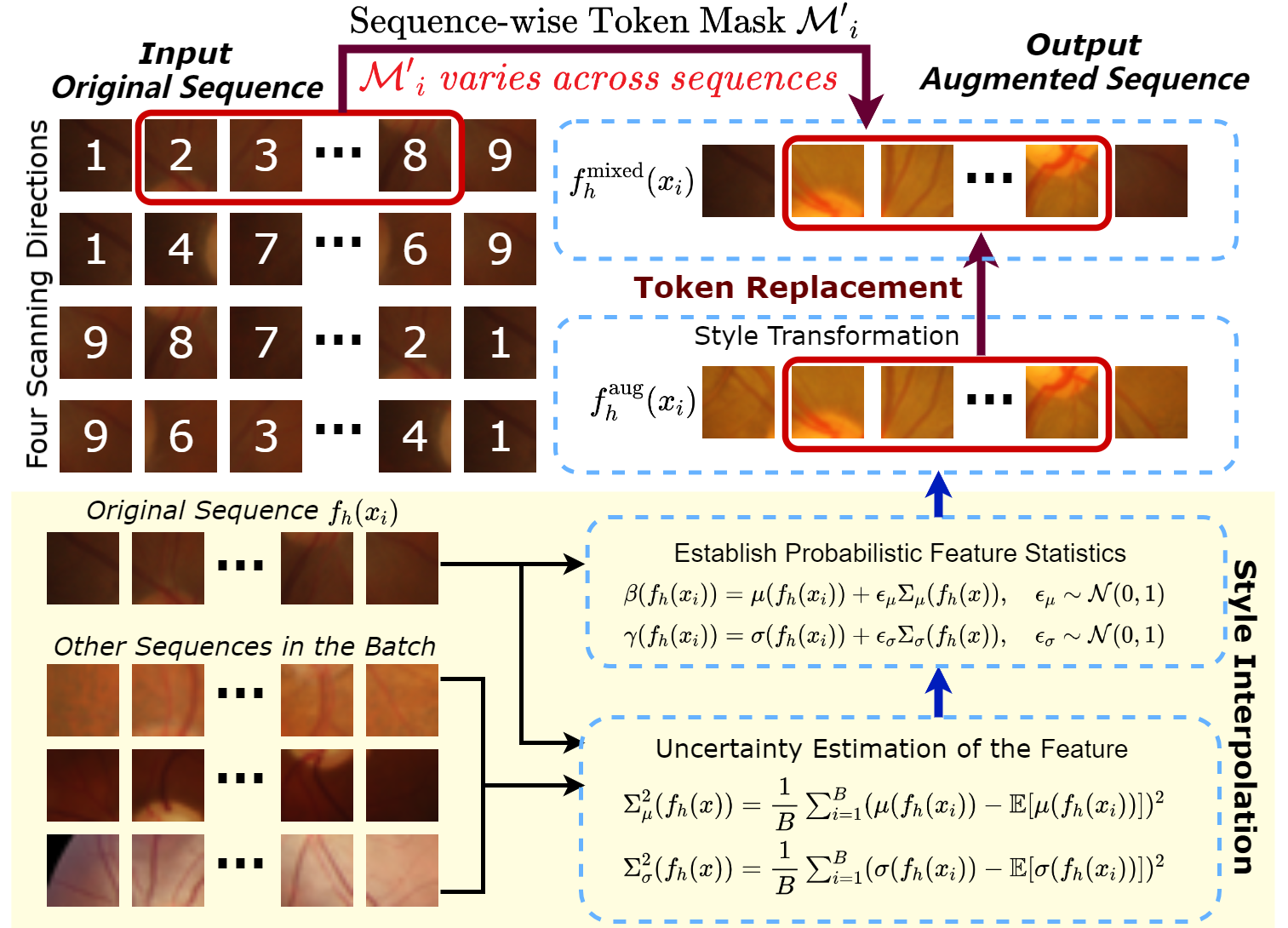}}
\caption{Structure of \textcolor{black}{LSA}. For each sequence, a unique mask is generated to enhance the diversity of the augmentation. 
}
\label{LOCALvis}
\end{figure}

\subsubsection{LSA: Local Sequence-wise Style Transformation Augmentation}
{Recognizing the distinctive characteristics of selective scanning mechanism within Mamba, we modify VSS Block in VMamba\cite{liu2024vmamba} to VSSDG Block shown in Fig. \ref{fig1}(c) by replacing SS2D with LSA to explicitly suppress domain-related features within the input-dependent matrixes of Mamba. 
LSA is designed for sequence-wise local style augmentation to further improve its generalization capabilities.
As shown in Fig. \ref{fig1}(d), LSA consists of five parts: scan expansion operation, selective SSMs module\cite{zhu2024vision}, style augmentation operation, sequence-wise mixup, and scan merging operation. 
Sequence-wise style augmentation is applied to each sequence in four directions. 
In consistent with preliminaries, $f_h(\cdot)$ denotes the features extracted from the middle layer of the SSM-based generalizable segmentation network. 
The detailed operation of LSA applied to the input is presented in Fig. \ref{LOCALvis} and described in text below.

Specifically, for the input representation $f_h(x) \in \mathbb{R}^{B\times D\times L}$ extracted from the selective SSMs module, where $x$ denotes a batch of images input to the SSM-based generalizable segmentation network, we compute \( \mu(x) \in \mathbb{R}^{B \times D} \) and \( \sigma(x) \in \mathbb{R}^{B \times D} \) as the mean and standard deviation of features computed channel-wise for each instance in a batch respectively. 
Let $f_h(x_i) \in \mathbb{R}^{D \times L}$ denote the input for the $i$-th sample in the batch, we define:
\begin{equation}
\begin{aligned}
    \mu(f_{h}(x_i)) &= \frac{1}{L}\textstyle \sum_{l=1}^{L} f_h(x_i)^l,\\
    \sigma(f_{h}(x_i)) &= \sqrt{\frac{1}{L}\textstyle \sum_{l=1}^{L} [f_h(x_i)^l - \mu(f_{h}(x_i))]^2 + \epsilon},
\end{aligned}
\end{equation}
where $\epsilon$ is a small constant for numerical stability. Next, we perform the following equations for uncertainty estimation:
\begin{equation}
\begin{aligned}
    \Sigma_{\mu}^{2}(f_{h}(x)) &= \frac{1}{B} \sum_{i=1}^{B} (\mu(f_{h}(x_i)) - \mathbb{E}[\mu(f_{h}(x_i))])^{2},\\
    \Sigma_{\sigma}^{2}(f_{h}(x)) &= \frac{1}{B} \sum_{i=1}^{B} (\sigma(f_{h}(x_i)) - \mathbb{E}[\sigma(f_{h}(x_i))])^{2},
\end{aligned}
\end{equation}
where \(\Sigma_{\mu}^{2}(f_{h}(x)) \in \mathbb{R}^{D}\) and \(\Sigma_{\sigma}^{2}(f_{h}(x)) \in \mathbb{R}^{D}\) are the uncertainty estimation of feature mean $\mu$ and feature standard deviation $\sigma$, \( \mathbb{E}_{b}(\cdot) \) represents the mathematical expectation over a mini-batch $b$. Then we establish a Gaussian distribution for feature statistics based on this estimation:
\begin{equation}
\begin{aligned}
    \beta(f_{h}(x_i)) &= \mu(f_{h}(x_i)) + \epsilon_{\mu} \Sigma_{\mu}(f_{h}(x)), \quad \epsilon_{\mu} \sim \mathcal{N}(0, 1),\\
    \gamma(f_{h}(x_i)) &= \sigma(f_{h}(x_i)) + \epsilon_{\sigma} \Sigma_{\sigma}(f_{h}(x)), \quad \epsilon_{\sigma} \sim \mathcal{N}(0, 1).
\end{aligned}
\end{equation}

Next, we replace the feature statistics to achieve style transformation. The input feature \( f_h(x_i) \) is converted to a transformed feature \( f_{h}^{\text{aug}}(x_i) \) using the following equation:
\begin{equation}
\begin{aligned}
    f_{h}^{\text{aug}}(x_i) = \beta(f_{h}(x_i)) + \left(\frac{f_{h}(x_i) - \mu(f_{h}(x_i))}{\sigma(f_{h}(x_i))}\right) \cdot \gamma(f_{h}(x_i)).
\end{aligned}
\end{equation}

\textcolor{black}{Unlike pervious DG methods that mainly perturb the entire feature maps at the channel level\cite{zhou2021mixstyle,guo2023aloft}, LSA incorporates a random-mask token selection scheme to perturb the regions of input-dependent matrics within Mamba.} 
We introduce the mask \( \mathcal{M'}_i \in \mathbb{R}^{1 \times L} \), which is defined as follows:
\begin{equation}
\begin{aligned}
\mathcal{M'}_{i,j} = \mathbb{I} \left( 0 \leq \left\lfloor \frac{j - j_{\text{start}}}{L} \right\rfloor < p \right), \quad j \in \{1, 2, ..., L\},
\end{aligned}
\label{mask2_local}
\end{equation}

The mask \( \mathcal{M'}_{i,j} \) is set to 1 for a proportion \( p \) of the indices, starting from a randomly chosen index \( j_{\text{start}} \), and 0 otherwise. Here, \( i \) represents the index of an element in the input tensor. 
The mask \( \mathcal{M'}_i \in \mathbb{R}^{1 \times L} \) is then expanded to \( D \times L \) dimensions, allowing it to blend element-wise with both \( f_h(x_i) \in \mathbb{R}^{D \times L} \) and \( f_h^{\text{aug}}(x_i) \in \mathbb{R}^{D \times L} \). 
Using broadcasting, the same \( L \)-dimensional mask is applied across all \( D \)-dimensions. Finally, the mixed tensor \( f_h^{\text{mixed}}(x_i) \) is computed by combining \( f^{\text{aug}}_h(x_i) \) and \( f_h(x_i) \) using the mask \( \mathcal{M'}_{i} \):

\begin{equation}
\begin{aligned}
    f_h^{\text{mixed}}(x_i) = f_h^{\text{aug}}(x_i) \odot \mathcal{M'}_{i} + f_h(x_i) \odot (1 - \mathcal{M'}_{i}).
\end{aligned}
\end{equation}

To preserve token dependencies, we choose continuous sequence perturbation over randomly selecting a proportion of pixels. 
Continuous sequences help maintain the structural integrity of features, ensuring that important semantic relationships are not disrupted by isolated perturbations.

Considering the structure of Mamba, our module employs a variety of augmentations to model uncertainty. \textcolor{black}{These augmentations synthesize feature statistics across sequences arranged in random continuous sub-sequences at the local level.}
Specifically, we hypothesize that each feature statistic follows a multivariate Gaussian distribution for modeling the diverse potential shifts. 
The generated feature statistics, with their diverse distribution possibilities, enable the models to achieve enhanced robustness against different domain shifts.

\subsubsection{Computational Burden}{In our GVA module, we introduced a learnable lightweight network with a parameter size of only $2.58 \times 10^{-4} \, \text{M}$. LSA introduces no additional parameters, involving only a few matrix operations during training, thereby maintaining a linear complexity similar to VM-Unet\cite{ruan2024vm}. 
Note that in the inference stage, GVA and LSA will be discarded, so there is no impact on computational burden. In terms of training time, after incorporating the GVA and LSA modules, the time per iteration \textcolor{black}{with the batch size of 8} on the Fundus dataset increased by only $0.004$ seconds ($0.342s \textit{ vs. } 0.338s$) compared to the baseline\cite{ruan2024vm}. This negligible increase in computational cost is inconsequential and could be disregarded. }

\subsection{Semantic Consistency Training}
To enhance the robustness and generalization capability of our segmentation model, we propose a semantic consistency training component. 
This component leverages predictions from both a base segmentation network and a modified network that includes GVA to enforce consistency in the predictions. 
The original input image $x_{p}^{k}$ is firstly processed through GVA to obtain $x_{p,E}^{k}$, which performs global augmentations. 

Both $x_{p}^{k}$ and $x_{p,E}^{k}$ are passed through the segmentation network. 
We define $\hat{y}_p^k$ as the predicted segmentation mask of $x_p^k$ and $\hat{y}_{p, E}^k$ as the predicting segmentation mask of $x_{p,E}^k$. We utilize the unified cross-entropy (CE) loss and Dice loss as our segmentation loss to optimize the model. The CE and Dice loss on the original image $x_p^k$ are formulated as:

\begin{equation}
\begin{aligned}
&\mathcal{L}_{\text{ce,O}}^{k} =\frac{1}{n_k} \sum_{p=1}^{n_k}\Big(y_p^k\log\hat{y}_p^k + (1 - y_p^k)\log(1 - \hat{y}_p^k)\Big), \\
&\mathcal{L}_{\text{dice,O}}^{k} = 1 - \frac{2\sum_{p=1}^{n_k} \hat{y}_p^k y_p^k}{\sum_{p=1}^{n_k}(\hat{y}_p^k + y_p^k + \epsilon)},
\end{aligned}
\end{equation}
where $y_p^k$ serves as the shared ground truth of $x_p^k$ and $x_{p,E}^{k}$, $\mathcal{L}_{\text{ce,O}}^{k}$ and $\mathcal{L}_{\text{dice,O}}^{k}$ are the CE loss and Dice loss measuring the difference between the ground truth and the predicted segmentation mask of original image respectively. 
The CE loss $\mathcal{L}_{\text{ce,E}}^{k}$ and Dice loss $\mathcal{L}_{\text{dice,E}}^{k}$ on $x_{p,E}^{k}$ are similar as above. Segmentation losses on $x_p^k$ and $x_{p,E}^{k}$ can be written as:

\begin{equation}
\begin{aligned}
\mathcal{L}_{\text{seg,O}}^{k} &= \mathcal{L}_{\text{ce,O}}^{k} + \mathcal{L}_{\text{dice,O}}^{k},\\
\mathcal{L}_{\text{seg,E}}^{k} &= \mathcal{L}_{\text{ce,E}}^{k} + \mathcal{L}_{\text{dice,E}}^{k}.
\end{aligned}
\end{equation}

To combat domain shifts, we propose a semantic consistency loss in our method. 
We force the segmentation model to predict consistent segmentation results from $x_p^k$ and $x_{p,E}^k$. 
So that the segmentation model can be less sensitive to domain shifts. We design a loss term to minimize the Mean Square Error between predictions $\hat{y}_p^k$ and $\hat{y}_{p, E}^k$. Our semantic consistency loss is written as:

\begin{equation}
\begin{aligned}
&\mathcal{L}_{\text{consist}}^k = \frac{1}{N} \sum_{p=1}^{N}\Big(\hat{y}_p^k - \hat{y}_{p, E}^k\Big)^2.
\end{aligned}
\end{equation}

Overall, we can formulate our framework as a multi-task learning paradigm. The total training loss is composed of several components designed to optimize both the segmentation accuracy and consistency between the two predictions. The loss function is defined as follows:
\begin{equation}
   \mathcal{L}_{\text{total}} = \frac{1}{K} \sum_{k=1}^{K}\Big(\mathcal{L}_{\text{seg,O}}^{k} +\mathcal{L}_{\text{seg,E}}^{k} + \lambda \mathcal{L}_{\text{consist}}^k\Big),
    \label{eq:total_loss}
\end{equation}
where $K$ represents the number of source domains and \( \lambda \) is the hyperparameter that balances the weights of the segmentation loss and consistency loss. 
The consistent training component is to promote prediction consistency between globally-augmented images and original images. 
This strategy closely integrates global appearance variation augmentation with local sequence-wise style transformation augmentation. By consistent training, GVA and LSA work in harmony to alleviate the accumulation of domain-specific features in the input-dependent matrices of Mamba during the recurrent process when the model encounters limited domains of medical image datasets, thereby minimizing reliance on domain-specific features.

\begin{table}
\centering
    \caption{\textcolor{black}{Statistics of Fundus and Prostate datasets used for our study.}}
    \label{tab:dataset}
    \renewcommand\arraystretch{1.05}
    \setlength{\tabcolsep}{0.7mm}{
    \begin{threeparttable}
    \textcolor{black}{\begin{tabular}{c||cccc}
        \toprule[1pt]
        \textcolor{black}{Task} & \textcolor{black}{Domain} & \textcolor{black}{Dataset} & \textcolor{black}{Cases in Each Domain} & \textcolor{black}{Total} \\ \hline
        \multirow{4}{*}{\textcolor{black}{Fundus}} & \textcolor{black}{1} & \textcolor{black}{Drishti-GS\cite{sivaswamy2015comprehensive}} & \textcolor{black}{50 / 51} & \multirow{4}{*}{\textcolor{black}{789 / 281\tnote{*}}} \\
        & \textcolor{black}{2} & \textcolor{black}{RIM-ONE-R3\cite{fumero2011rim}} & \textcolor{black}{99 / 60} & \\
        & \textcolor{black}{3} & \textcolor{black}{REFUGE-train\cite{orlando2020refuge}} & \textcolor{black}{320 / 80} & \\
        & \textcolor{black}{4} & \textcolor{black}{REFUGE-val\cite{orlando2020refuge}} & \textcolor{black}{320 / 80} & \\
        \hline
        \multirow{6}{*}{\textcolor{black}{Prostate\tnote{\#}}} & \textcolor{black}{1} & \textcolor{black}{NCI-ISBI13\cite{bloch2015nci}} & \textcolor{black}{30} & \multirow{6}{*}{\textcolor{black}{116}} \\
        & \textcolor{black}{2} & \textcolor{black}{NCI-ISBI13\cite{bloch2015nci}} & \textcolor{black}{30} & \\
        & \textcolor{black}{3} & \textcolor{black}{I2CVB\cite{lemaitre2015computer}} & \textcolor{black}{19} & \\
        & \textcolor{black}{4} & \textcolor{black}{PROMISE12\cite{litjens2014evaluation}} & \textcolor{black}{13} & \\
        & \textcolor{black}{5} & \textcolor{black}{PROMISE12\cite{litjens2014evaluation}} & \textcolor{black}{12} & \\
        & \textcolor{black}{6} & \textcolor{black}{PROMISE12\cite{litjens2014evaluation}} & \textcolor{black}{12} & \\
        \bottomrule[1pt]
    \end{tabular}}
    \begin{tablenotes}
    \footnotesize
    \item[\textcolor{black}{*}] \textcolor{black}{Data split (training / test cases) was provided by \cite{14}.}
    \item[\textcolor{black}{\#}] \textcolor{black}{NCI-ISBI13 and PROMISE12 actually include multiple data sources.}
    \end{tablenotes}
    \end{threeparttable}
    }
\end{table}

\begin{table*} [!t]
\centering
\caption{Dice coefficient [\%] and ASD [voxel] of different methods on \textbf{Fundus} dataset. We run our experiment three times and report the mean performance. Top results are highlighted in \textbf{bold}.}
\scriptsize
    \renewcommand\arraystretch{1.05}
\label{tab:results_fundusdice}
    \setlength{\tabcolsep}{1.2mm}{
        \begin{tabular}{c||cccc|c||cccc|c}
            \toprule[1.0pt]
            Task & \multicolumn{4}{c|}{Optic Cup / Disc Segmentation (Dice coefficient $\uparrow$)} & \multirow{2}{*}{Avg.} & \multicolumn{4}{c|}{Optic Cup / Disc Segmentation (ASD $\downarrow$)} & \multirow{2}{*}{Avg.} \\
            \cmidrule[0.3pt]{1-5}\cmidrule[0.3pt]{7-10}
            Unseen Site & Domain 1 & Domain 2 & Domain 3 & Domain 4&& Domain 1 & Domain 2 & Domain 3 & Domain 4 & \\
            \midrule[0.3pt]
            \text{\tiny 19'}JiGen \cite{carlucci2019domain} & 82.45 / 95.03 & 77.05 / 87.25 & 87.01 / 94.94 & 80.88 / 91.34&86.99 & 18.57 / 9.43 &17.29 / 19.53&9.15 / 6.99&15.84 / 12.14&13.62 \\
            \text{\tiny 20'}BigAug \cite{zhang2020generalizing} & 77.68 / 93.32 & 75.56 / 87.54 & 83.33 / 92.68 & 81.63 / 92.20&85.49 & 22.61 / 12.53&17.95 / 17.64&11.48 / 10.33&11.57 / 9.36&14.18 \\
            \text{\tiny 20'}DoFE \cite{14} & 83.59 / 95.59 & 80.00 / 89.37 & 86.66 / 91.98 & 87.04 / 93.32& 88.44& 16.90 / 7.68&13.87 / 16.59&9.59 / 11.19&7.24 / 7.53&10.75 \\
            \text{\tiny 21'}FedDG \cite{liu2021feddg} & 84.13 / 95.37 & 71.88 / 87.52 & 83.94 / 93.37 & 85.51 / 94.50 &87.03& 18.57 / 7.69&15.87 / 16.93&11.09 / 7.28&10.23 / 7.51&11.90 \\
            \text{\tiny 22'}RAM-DSIR \cite{13} & 85.48 / 95.75 & 78.82 / 89.43 & \textbf{87.44} / 94.67 & 85.84 / 94.10&88.94& 16.05 / 7.12&14.01 / 13.86&\textbf{9.02} / 7.11&8.29 / 7.06&10.32 \\
            \text{\tiny 22'}DCAC \cite{hu2022domain} & 81.43 / 96.54 & 77.72 / 87.45 & 86.80 / 94.28 & 87.68 / 95.40 &88.47&  19.20 / \textbf{6.35}&17.15 / 18.28&9.14 / 8.11&7.12 / \textbf{5.20}&11.32 \\
            \text{\tiny 24'}WT-PSE \cite{chen2024learning} & 84.55 / 95.45 & 77.00 / 87.94 & 86.10 / \textbf{95.26} & 84.77 / 93.68 &88.10& 16.34 / 8.07&16.12 / 16.99&10.07 / 9.04 &9.60 / 7.81&11.76 \\
            \text{\tiny 23'}SAM-Med2D \cite{cheng2023sam}&84.78 / 92.14& \textbf{80.81} / 87.63& 78.02 / 90.83&78.62 / 90.02&85.36&15.38 / 8.89&\textbf{12.04} / 13.72&14.64 / 8.51&11.47 / 9.09&11.72\\
            \textcolor{black}{\text{\tiny 23'}Med-SA \cite{wu2023medical}}&\textcolor{black}{81.64 / 93.24}&\textcolor{black}{73.74 / 85.33}&\textcolor{black}{76.57 / 93.87}&\textcolor{black}{75.13 / 85.52}&\textcolor{black}{83.13}&\textcolor{black}{13.49 / 7.91}&\textcolor{black}{16.77 / 20.63}&\textcolor{black}{10.81 / \textbf{4.44}}&\textcolor{black}{19.08 / 18.10}&\textcolor{black}{13.90}\\
            \midrule[0.3pt]
            Baseline\cite{ruan2024vm} & 82.48 / 95.42& 76.22 / 88.57 & 83.33 / 91.76& 84.20 / 93.28& 86.90&16.63 / 9.47& 15.23 / 15.20&11.96 / 10.68&10.59 / 7.74&12.19\\
             \rowcolor{lightgray}
             \textbf{Mamba-Sea} (Ours) &\textbf{85.93} / \textbf{96.87}&79.56 / \textbf{90.61}&86.45 / 93.95&\textbf{88.18} / \textbf{95.87}&\textbf{89.68}&\textbf{14.30} / 7.49&12.72 / \textbf{13.18}&9.90 / 8.26&\textbf{7.04} / 5.70&\textbf{9.82}\\
            \bottomrule[0.8pt]
        \end{tabular}
    }

\end{table*}

\section{Experiments}
\subsection{Datasets}
In line with the work of \cite{13, liu2021feddg}, we evaluate our method on two public domain generalization (DG) medical image segmentation datasets: \textbf{Fundus}\cite{14} and  \textbf{Prostate}\cite{liu2020shape}. 

\begin{table*} [!t]
\centering
\caption{Dice coefficient [\%] and ASD [voxel] of different methods on \textbf{Prostate} dataset. Top results are highlighted in \textbf{bold}.}
\scriptsize
    \renewcommand\arraystretch{1.05}
\label{tab:results_prodice}
    \setlength{\tabcolsep}{0.55mm}{
        \begin{tabular}{c||cccccc|c|cccccc|c}
            \toprule[1.0pt]
            Task & \multicolumn{6}{c|}{Prostate Segmentation (Dice coefficient $\uparrow$)}& \multirow{2}{*}{Avg.}& \multicolumn{6}{c|}{Prostate Segmentation (ASD $\downarrow$)}& \multirow{2}{*}{Avg.}\\
            \cmidrule[0.3pt]{1-7}\cmidrule[0.3pt]{9-14}
            Unseen Site & Domain 1 & Domain 2 & Domain 3 & Domain 4& Domain 5 & Domain 6&&Domain 1 & Domain 2 & Domain 3 & Domain 4& Domain 5 & Domain 6& \\
            \midrule[0.3pt]
            \text{\tiny 19'}JiGen \cite{carlucci2019domain}&85.45&89.26&85.92&87.45&86.18&83.08&86.22&1.11&1.81&2.61&1.66&1.71&2.43&1.89\\
            \text{\tiny 20'}BigAug \cite{zhang2020generalizing}&85.73&89.34&84.49&88.02&81.95&87.63&86.19&1.13&1.78&4.01&1.25&1.92&1.89&2.00\\
            \text{\tiny 20'}DoFE \cite{14} &89.64&87.56&85.08&89.06&86.15&87.03&87.42&\textbf{0.92}&1.49&2.74&1.46&1.89&1.53&1.68  \\
            \text{\tiny 21'}FedDG \cite{liu2021feddg} &90.19&87.17&85.26&88.23&83.02&90.47&87.39&1.30&1.67&2.36&1.37&2.19&1.94&1.81  \\
            \text{\tiny 22'}RAM-DSIR \cite{13} &87.56&90.20&86.92&88.72&87.17&87.93&88.08&1.04&0.81&2.23&1.16&1.81&1.15&1.37  \\
            \text{\tiny 22'}DCAC \cite{hu2022domain} & 91.76&90.51&86.30&89.13&83.39&90.56&88.61&0.98&0.89&1.77&1.53&2.46&\textbf{0.85}&1.41 \\
            \textcolor{black}{\text{\tiny 24'}WT-PSE \cite{chen2024learning}}&\textcolor{black}{85.70}&\textcolor{black}{85.27}&\textcolor{black}{86.91}&\textcolor{black}{88.09}&\textcolor{black}{85.73}&\textcolor{black}{86.08}&\textcolor{black}{86.30}&\textcolor{black}{2.11}&\textcolor{black}{3.04}&\textcolor{black}{2.30}&\textcolor{black}{1.91}&\textcolor{black}{2.56}&\textcolor{black}{1.61}&\textcolor{black}{2.26}\\
            \text{\tiny 23'}SAM-Med2D \cite{cheng2023sam}&86.50&87.19&\textbf{87.29}&85.78&\textbf{87.65}&87.48&86.98&2.01&2.38&1.96&2.14&\textbf{1.15}&1.72&1.89\\
             \textcolor{black}{\text{\tiny 23'}Med-SA \cite{wu2023medical}}&\textcolor{black}{82.06}&\textcolor{black}{85.97}&\textcolor{black}{82.59}&\textcolor{black}{86.66}&\textcolor{black}{83.08}&\textcolor{black}{86.56}&\textcolor{black}{84.49}&\textcolor{black}{3.04}&\textcolor{black}{1.97}&\textcolor{black}{3.68}&\textcolor{black}{4.05}&\textcolor{black}{4.49}&\textcolor{black}{2.15}&\textcolor{black}{3.23}\\
            \midrule[0.3pt]
            Baseline\cite{ruan2024vm} &91.02&91.41&81.71&89.18&81.14&90.91&87.56&1.28&1.45&2.15&2.72&2.65&2.30&2.09 \\
             \rowcolor{lightgray}
              \textbf{Mamba-Sea} (Ours) &\textbf{92.96}&\textbf{93.37}&84.40&\textbf{91.25}&85.91&\textbf{92.25}&\textbf{90.02}&1.08&\textbf{0.69}&\textbf{1.74}&\textbf{1.04}&1.87&1.13&\textbf{1.26} \\
            \bottomrule[0.8pt]
        \end{tabular}
    }
\end{table*}

The Fundus dataset contains $789$ cases for training and $281$ cases for test, which are collected from four public fundus image datasets and have inconsistent statistical characteristics. 
For preprocessing, we follow the previous work\cite{liu2020shape,14}, which involves center-cropping disc regions to an $800 \times 800$ bounding box for all images in the Fundus dataset. Subsequently, we randomly resize and crop a $256 \times 256$ region from each cropped image to serve as input for the network. \textcolor{black}{The Prostate dataset comprises $116$ T2-weighted MRI prostate images collected from six sources for segmentation}. All images are cropped to the $3$D prostate region, and the $2$D axial slices are resized to $384 \times 384$. For model training, we feed these 2D slices into our model. We normalize the intensity values of the data on both datasets individually to the range of $[-1, 1]$. \textcolor{black}{The
statistics of two datasets are summarized in Tab. \ref{tab:dataset}.}

\subsection{Implementation Details}
Mamba-Sea is a DG segmentation model that effectively integrates global enhancements with local sequence-wise augmentation. The experiment is implemented using the PyTorch framework on the Nvidia Tesla V$100$-SXM$2$ GPU with $32$ GB of memory. 
Following VM-UNet, we initialize the weights of both the encoder and decoder with those of VMamba-S\cite{liu2024vmamba} pre-trained on ImageNet \cite{deng2009imagenet}. 
We train our model for $20000$ iterations on Fundus and Prostate datasets, using a batch size of $8$ and $6$ for Fundus and Prostate datasets respectively. The model optimization is carried out using the Adamw optimizer with an initial learning rate of $0.0003$, and we employ a polynomial decay rule to stabilize the learning process. We set $\tau$ in Eq. \eqref{mask_global} to $0.4$ for Fundus dataset and $0.03$ for Prostate dataset. Finally, we set the value of p to $0.75$ in Eq. \eqref{mask2_local} and the values of \(\lambda\) to $0.1$ in Eq. \eqref{eq:total_loss}.

All of the implementations on datasets follow previous methods\cite{liu2020shape,13,14}. In the Fundus dataset, where each domain is already split into training and testing sets, we train our model on the source domains' training sets and evaluate it on the target domains' testing sets. During testing, we first resize the $800 \times 800$ test images to $256 \times 256$, obtain the segmentation masks, and then resize these masks back to $800 \times 800$ for evaluation metric computation. 
For the Prostate dataset, since the original images are $3$D volumes, we first generate $2$D predictions, concatenate these predictions for each $3$D sample, and compute the evaluation metrics on the resulting $3$D predictions. Additionally, we skip $2$D slices that do not contain any prostate regions during testing.

For evaluation, we employ the Dice coefficient and Average Surface Distance (ASD) as quantitative metrics to assess segmentation performance. A higher Dice coefficient indicates superior overall segmentation quality, while a lower ASD reflects more accurate surface delineation. Together, these metrics provide a comprehensive evaluation of both the volumetric accuracy and boundary precision of the segmentation models.

\subsection{Comparison with SOTA methods}
In our experiments, we adopt the leave-one-domain-out strategy commonly used in domain generalization literature\cite{13}. This approach involves training on \(K\) source domains and testing on the remaining target domain, resulting in a total of \(K + 1\) domains. Consequently, for the Fundus dataset, we have four distinct tasks, and for the Prostate dataset, we have six distinct tasks. This strategy ensures a comprehensive evaluation across different domains, allowing us to assess the generalizability of our model more effectively.

\begin{figure*}[tp]  
    \centering   
    \centerline{\includegraphics[width=1\linewidth]{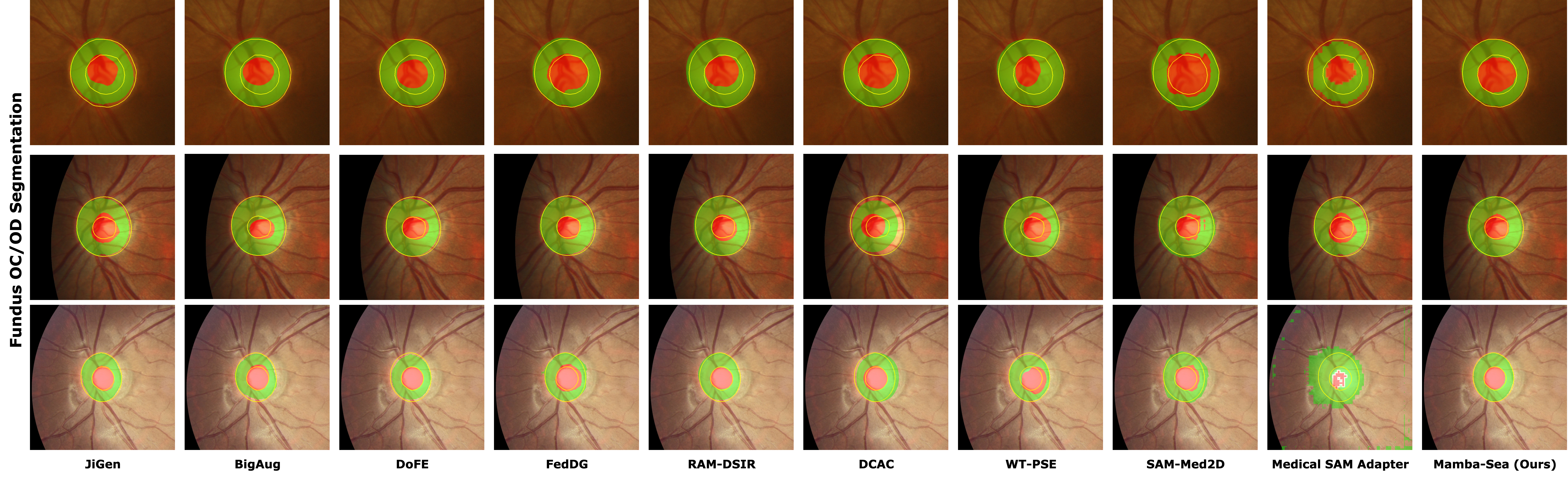}}
    \caption{
    Visualization on segmentation results of different methods on \textbf{Fundus} dataset. The yellow contours indicate the boundaries of ground truths while the semi-transparent overlays are predictions.
    } 
    \label{mask1}
\end{figure*}

\begin{figure*}[tp]  
    \centering   
    \centerline{\includegraphics[width=1\linewidth]{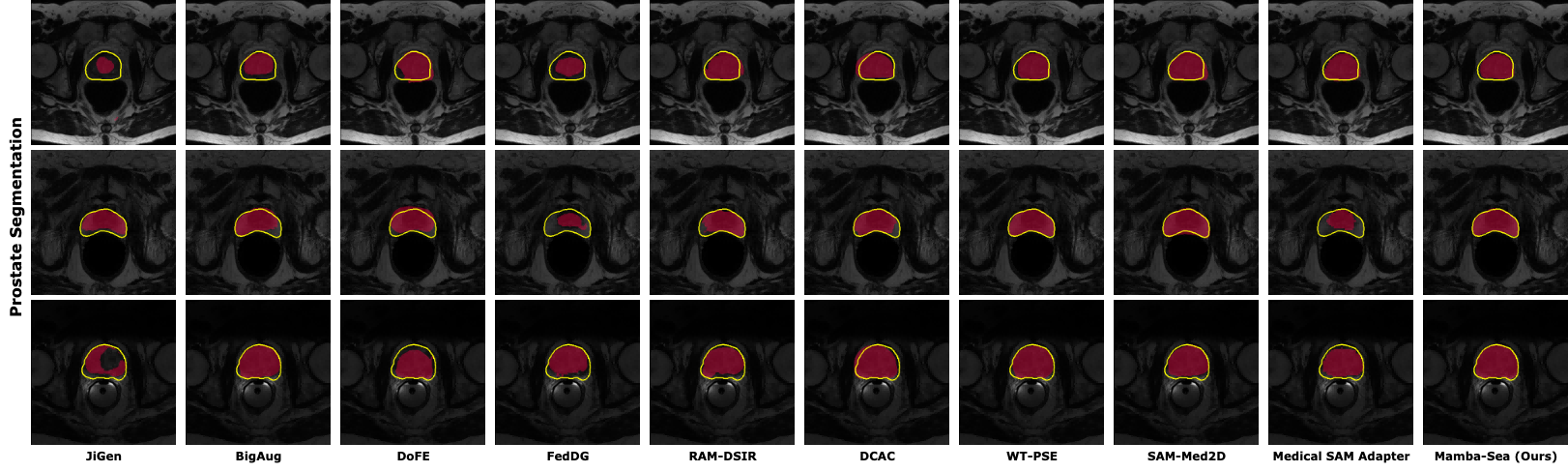}}
    \caption{
    Visualization on segmentation results of different methods on \textbf{Prostate} dataset. The yellow contours indicate the boundaries of ground truths while the semi-transparent overlays are predictions.
    } 
    \label{mask2}
\end{figure*}

We compare the results of our method with seven recent SOTA DG methods and a representative foundation model on Fundus dataset. 
JiGen\cite{carlucci2019domain} is an effective self-supervised DG method that enhances model regularization through the use of jigsaw puzzles. BigAug\cite{zhang2020generalizing} is a deep stacked transformation approach which used four image-based characteristics for data augmentation. DoFE\cite{14} employs Domain-oriented Feature Embedding to learn domain prior information from multi-source domains. 
FedDG\cite{liu2021feddg} is a meta-learning-based generalizable medical image segmentation method with a boundary-oriented episodic learning scheme. 
RAM-DSIR\cite{13} uses per-domain batch normalization by comparing target domain distributions with stored statistics from other domains. 
DCAC\cite{hu2022domain} introduces a multi-source model with domain and content-adaptive convolution modules for segmentation across different modalities.  
WT-PSE\cite{chen2024learning} uses a whitening transform-based shape regularization and knowledge distillation to enhance cross-domain performance under domain shift conditions. 

\textcolor{black}{In addition to selecting recent DG methods specifically designed to address domain shifts in medical image segmentation, we incorporate two SAM-based approaches in our investigation. This decision is motivated by the emergence of large-scale models and our objective to evaluate their performance under conditions where the testing set exhibits distributional shifts relative to the training set. The inclusion of these SAM-based methods allows us to systematically assess their capability in handling domain generalization challenges.}
SAM-Med2D\cite{cheng2023sam} leverages a high-parameter model and incorporates extensive additional training data to ensure robust performance across a wide range of medical image segmentation tasks. 
\textcolor{black}{We do not train SAM-Med2D; instead, we directly provided box prompts on the testing set for the model to make predictions. 
Med-SA \cite{wu2023medical} enhances SAM's segmentation capabilities for medical images by fine-tuning the SAM model \cite{kirillov2023segment} using LoRA\cite{hu2021lora}. For Med-SA, we employed 3-point click prompts for fine-tuning in the source domains.}
We train the basic VM-Unet\cite{ruan2024vm} by simply aggregating all source domain images as our baseline segmentation model.

\begin{table*}[]
    \centering
    \caption{Ablation Study of key components in our framework on \textbf{Fundus} and \textbf{Prostate} datasets. Using Dice coefficient [\%] as the evaluation metric to assess segmentation performance.}
    \label{tab:ab_fun}
        \renewcommand\arraystretch{1.05}
    \setlength{\tabcolsep}{0.75mm}{
    \begin{tabular}{cc||cccc|c||cccccc|c}
        \toprule[1pt]
        \multicolumn{2}{c||}{Task} & \multicolumn{4}{c|}{Optic Cup / Disc Segmentation} & \multirow{2}{*}{\textbf{Avg.}}& \multicolumn{6}{c|}{Prostate Segmentation}& \multirow{2}{*}{\textbf{Avg.}} \\
        \cmidrule[0.3pt]{1-6} \cmidrule[0.3pt]{8-13}
        \textcolor{black}{GVA} &\textcolor{black}{LSA}  & Domain 1 & Domain 2 & Domain 3 & Domain 4&& Domain 1 & Domain 2 & Domain 3 & Domain 4& Domain 5 & Domain 6   \\
        \cmidrule[0.3pt]{1-7} \cmidrule[0.3pt]{8-14}
        \XSolidBrush & \XSolidBrush  &82.48 / 95.42&76.22 / 88.57&83.33 / 91.76&84.20 / 93.28&86.90&91.02&91.41&81.71&89.18&81.14&90.91&87.56 \\
        \CheckmarkBold & \XSolidBrush &85.81 / 96.63&76.55 / 88.40&\textbf{87.64} / 93.41&80.58 / 93.67&87.84&91.29&91.52&82.38&89.33&80.85&90.95&87.72 \\
        \XSolidBrush & \CheckmarkBold & 82.78 / 96.15 & \textbf{80.03} / 89.82&84.81 / 92.70&87.66 / 95.26&88.65&90.96&91.44&82.50&89.04&\textbf{87.59}&91.19&88.79 \\
        \CheckmarkBold & \CheckmarkBold &85.88 / 96.83&78.56 / 90.16&86.41 / 93.74&83.10 / 94.81&88.69&91.63&92.24&84.13&89.07&87.11&91.43&89.27\\
        \cmidrule[0.3pt]{1-14}
        \rowcolor{lightgray}
        \multicolumn{2}{c||}{Mamba-Sea}&\textbf{85.93} / \textbf{96.87}&79.56 / \textbf{90.61}&86.45 / \textbf{93.95}&\textbf{88.18} / \textbf{95.87}&\textbf{89.68}&\textbf{92.96}&\textbf{93.37}&\textbf{84.40}&\textbf{91.25}&85.91&\textbf{92.25}&\textbf{90.02}\\
        \bottomrule[1pt]
    \end{tabular}
    }
\end{table*}

In Tab. \ref{tab:results_fundusdice}, we show Dice coefficient and ASD results of different domains on Fundus dataset. 
\textcolor{black}{We run our experiment three times and report the mean performance.} 
For fairness, we report the data as presented in pervious works. The results of Domain \( k \), where \( k \in \{1, 2, 3, 4\} \), are obtained from the model trained using images from the other three domains. 
Notably, even our baseline model, the VM-Unet\cite{ruan2024vm} based on Mamba, demonstrates exceptional performance, achieving results approaching FedDG\cite{liu2021feddg} (Dice coefficient $86.90\% \textit{ vs. } 87.03\%$; ASD $12.19$ voxel \textit{vs.} $11.90$ voxel). This observation underscores the potential of Mamba, particularly its sequential scanning approach, making it a valuable method for addressing domain shifts in medical image segmentation. 
However, all methods demonstrate improvements over the baseline to varying degrees, indicating that different regularization and generalization strategies can enhance the model’s ability to learn more robust feature representations. Compared to these methods, our method achieves higher average Dice coefficients and lower average ASD on the Fundus dataset. This success is attributed to our combined global and local augmentation strategies. Additionally, we adopt a collaborative training strategy to mitigate domain shifts. These components contribute to the superior performance of our method on the Fundus dataset. 
Compared to the baseline, our approach consistently improves performance across all unseen domain settings, with an average Dice coefficient increase of $2.78\%$ and an average ASD decrease of $2.37$ voxel (Dice coefficient $86.90\% \textit{ vs. } 89.68\%$; ASD $12.19$ voxel \textit{vs.} $9.82$ voxel). Moreover, our framework achieves SOTA performance compared with competitive methods.

To further demonstrate the effectiveness of our method, we present experimental results on the Prostate dataset in Tab. \ref{tab:results_prodice}. For the prostate segmentation task, our baseline model also performs well, even surpassing DoFE\cite{14} and FedDG\cite{liu2021feddg} on the evaluation metrics of Dice coefficient. 
Our framework achieves the highest Dice coefficient and ASD across most unseen domains, with an average Dice coefficient of $90.02\%$ and an ASD of $1.26$ voxel, both of which are superior to other DG methods. \textcolor{black}{Compared to the baseline, our method increases the overall Dice coefficient by $2.46\%$ and reduces the ASD by $40\%$ to $1.26$ voxel (Dice coefficient $87.56\% \textit{ vs. } 90.02\%$; ASD $2.09$ voxel \textit{vs.} $1.26$ voxel)}. In Fig. \ref{mask1} and Fig. \ref{mask2}, we visualize the segmentation results of our method alongside several approaches for medical images on the Fundus and Prostate datasets. Compared to other methods, our approach produces more accurate segmentations with smoother boundaries.

\subsection{Ablation Study and Further Analyses}
\subsubsection{Ablation Study of Framework's Components}
{\textcolor{black}{We perform comprehensive ablation studies to systematically assess the individual contributions of our proposed Global-to-Local augmentation modules and the consistent training strategy across both the Fundus and Prostate datasets. Firstly, we establish our baseline by removing all three key components: GVA, LSA, and the consistency learning strategy. This baseline configuration, which corresponds to the results presented in the first row of Tab. \ref{tab:ab_fun} and serves as baseline in Tab. \ref{tab:results_fundusdice} and \ref{tab:results_prodice}. 
Secondly, we investigate the impact of LSA removal while maintaining GVA and consistency constraints. In this configuration, images undergo globally augmentation and are subjected to consistency regularization with their original counterparts. 
Thirdly, we eliminate GVA and the consistency training strategy, allowing images to be processed solely through LSA for locally generalization. 
Finally, we examine the model's performance without consistency constraints while retaining both GVA and LSA. 
This final ablation setting provides crucial insights into the robustness of our proposed modules by quantifying the contribution of the consistency regularization component. 
From the experimental results reported in Tab. \ref{tab:ab_fun}, the absence of LSA leads to a significant performance degradation, with the Dice coefficient decreasing by $1.84\%$ on the Fundus dataset ($87.84\% \textit{ vs. } 89.68\%$) and $2.3\%$ on the Prostate dataset ($87.72\% \textit{ vs. } 90.02\%$), respectively.}
All augmentation modules and the consistent training strategy significantly improve performance, with the removal of either module resulting in a noticeable drop in accuracy. 
They contribute together to enhancing our model's generalization performance, implying that our model is more robust to domain shifts in medical image segmentation.}

\subsubsection{Further Analysis of GVA and LSA}
{
Mamba-Sea incorporates global perturbations through GVA and local style augmentation through LSA. These components simulate diverse domain shifts in sequence token dependencies, selectively perturbing and suppressing domain-specific features in salient tokens while preserving semantic content. 
The GVA and LSA modules are not simply combined; they are carefully designed based on a thorough analysis of the requirements of domain-generalized medical image segmentation in the context of Mamba.
To further validate the effectiveness of our modules, we conduct comparative experiments through two distinct strategies. 
First, we perform an ablation study by substituting GVA with Gamma Correction while maintaining both LSA and semantic consistency training components. The results are shown in Tab. \ref{tab:gvawithmamba}. This configuration allows for a direct comparison of different global augmentation strategies.
\begin{table}
\centering
    \caption{\textcolor{black}{Effects of different settings of GVA. Using Dice coefficient [\%] as the evaluation metric to assess performance.}}
        \renewcommand\arraystretch{1.05}
    \label{tab:gvawithmamba}
    \setlength{\tabcolsep}{2mm}{
    \textcolor{black}{
    \begin{tabular}{c||cc}
        \toprule[1pt]
        Model&Fundus&Prostate\\
        \cmidrule[0.3pt]{1-3}
        VM-UNet + LSA&88.65&88.79\\
           \midrule[1pt]
        + Gamma Correction (Gamma = 0.5)&88.72&89.01\\
        + Gamma Correction (Gamma = 2.0)&88.97&89.16\\
        + GVA (w/o gating machanism)&89.07&88.98\\
        + GVA&\textbf{89.68}&\textbf{90.02}\\
        \bottomrule[1pt]
    \end{tabular}
    }}
\end{table}
\begin{table}
\centering
    \caption{\textcolor{black}{Comparision with CNN-based or ViT-based DG methods and different inserted positions of LSA on Mamba. Using Dice coefficient [\%] as the evaluation metric to assess performance.}}
    \renewcommand\arraystretch{1.05}
    \label{tab:lsawithmamba}
    \setlength{\tabcolsep}{1mm}{
    \textcolor{black}{{\begin{tabular}{c|cc||c|cc}
        \toprule[1pt]
        Model&Fundus&Prostate&Model&Fundus&Prostate\\
        \cmidrule[0.3pt]{1-6}
        VM-UNet \cite{ruan2024vm}&86.90&87.56&VM-UNet \cite{ruan2024vm}&86.90&87.56\\
           \midrule[1pt]
        + CutMix \cite{yun2019cutmix}&87.91&87.86&\multirow{2}{*}{+ LSA (EN)}&\multirow{2}{*}{88.46}&\multirow{2}{*}{88.61}\\
         + MixStyle \cite{zhou2021mixstyle}&87.27&87.60&&\\
        + ALOFT-S \cite{guo2023aloft}&88.03&88.47&+ LSA (DE)&87.83&88.04\\
        + ALOFT-E \cite{guo2023aloft}&87.75&87.78&\multirow{2}{*}{+ LSA (EN+DE)}&\multirow{2}{*}{\textbf{88.65}}&\multirow{2}{*}{\textbf{88.79}}\\
        + LSA&\textbf{88.65}&\textbf{88.79}&&\\
        \bottomrule[1pt]
    \end{tabular}
}}}
\end{table}
Second, we benchmark LSA against SOTA DG methods implemented on the VM-UNet backbone. The comparison encompasses representative CNN-based approaches, including MixStyle \cite{zhou2021mixstyle} and CutMix \cite{yun2019cutmix}, as well as the ViT-based ALOFT \cite{guo2023aloft}. 
To highlight the distinctive advantages of our LSA module in these comparative experiments, we deliberately remove both the GVA module and the consistency training component, thereby isolating the performance contribution of our LSA implementation. 
Third, we conduct experiments applying LSA exclusively in the encoder section and separately in the decoder section of VM-Unet, while retaining original SS2D in other components. 
The results presented in Tab. \ref{tab:lsawithmamba} reveal that incorporating the modules in both encoder and decoder sections (EN + DE) with LSA yields the best performance, suggesting that modeling uncertainty across all training stages enhances the overall effectiveness of the method. 
\begin{figure}[!t]  
    \centering   
    \centerline{\includegraphics[width=0.9\linewidth]{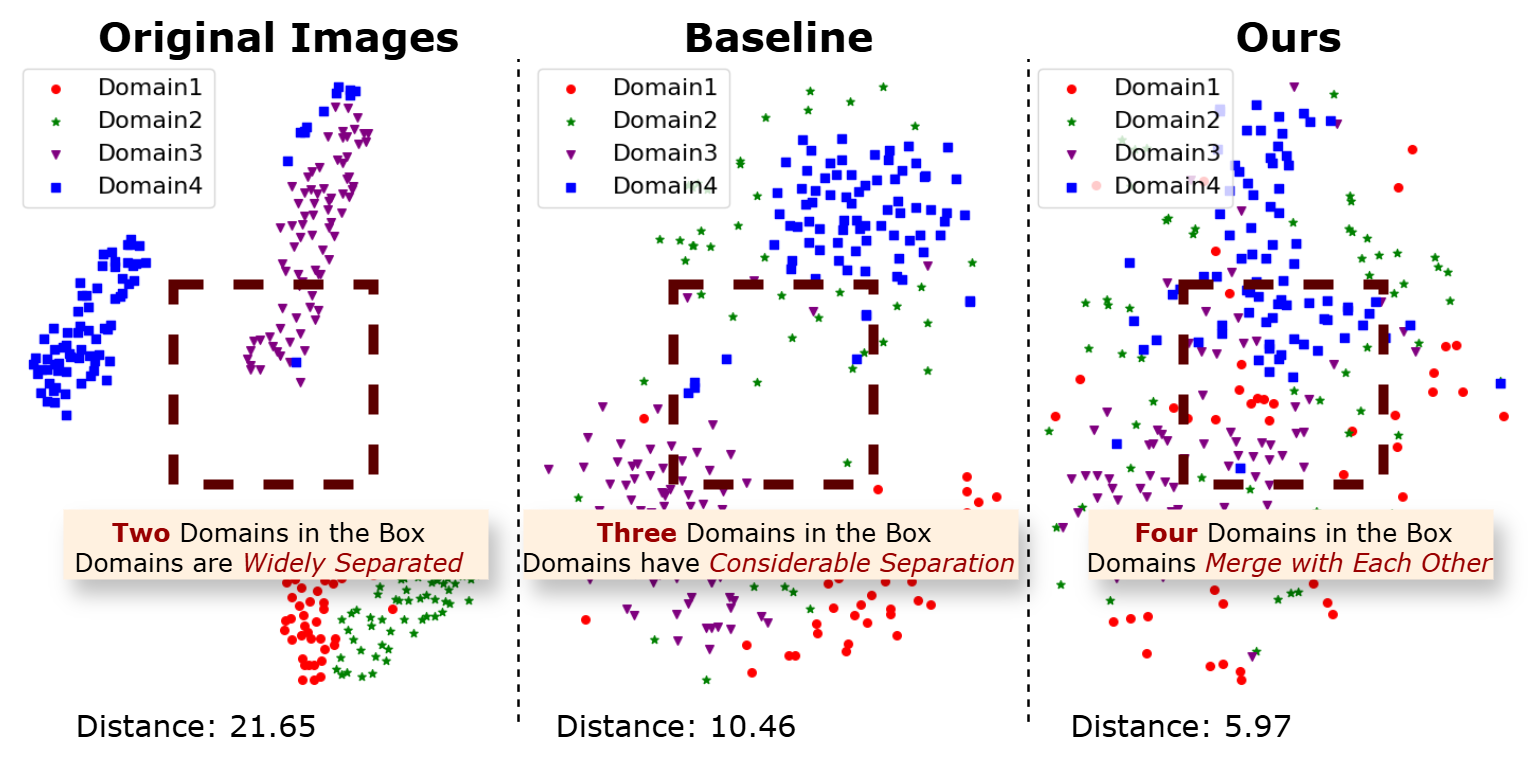}}
    \caption{
    t-SNE visualization\cite{van2008visualizing} of the features extracted from original images, after training with VM-UNet\cite{ruan2024vm} and Mamba-Sea on the \textbf{Fundus} dataset. Different colors represent different domains. The distance between domains is computed as the average Euclidean distance between centroids of each domain.
    } 
    \label{tsne}
    \vspace{-0.5cm}
\end{figure}

\subsubsection{Generalization Performance of our Framework}
{
We conduct a t-SNE visualization\cite{van2008visualizing} to analyse the features extracted from the original images, the features obtained after training with VM-UNet, and those extracted after training with our proposed framework on the Fundus dataset to analyse the generalization performance. 
As shown in the red box of Fig. \ref{tsne}, we observe that the distance between domains in the original images is large. 
After training with our baseline, the VM-UNet, \textcolor{black}{we find domains still has considerable separation with each other. This observation suggests that when processing medical image datasets with domain shifts, the input-dependent matrices in the Mamba architecture tend to accumulate domain-specific characteristics throughout the recurrent processing stages, thereby compromising the model's generalization capability across different domains.} 
Notably, when trained with Mamba-Sea, the distance is further minimized. This result further supports our hypothesis that the Mamba architecture is effective for DG in medical image segmentation. Mamba-Sea effectively explores the potential of Mamba, allowing the model to efficiently overcome domain shifts.
\textcolor{black}{To further validate the capability of our method in reducing domain gaps within the context of input-dependent matrices during the training process, we conduct additional quantitative experiments following the previous work \cite{guo2024start}. Specifically, we compare domain gaps across different methods using the Fundus dataset with VM-UNet, focusing on the feature maps generated by the final block in the decoder pathway. The experimental results demonstrate a $32\%$ reduction in the average domain gap ($0.333 
 \textit{ vs.} 0.225$) across the four domains, providing concrete evidence that Mamba-Sea effectively mitigates domain discrepancies in the input-dependent matrices for medical image segmentation.}
}

\begin{table}
\centering
    \caption{\textcolor{black}{Parameter sensitivity analysis by varying parameter $p$ and $\lambda$ on \textbf{Fundus} and \textbf{Prostate} datasets. Using Dice coefficient [\%] as the evaluation metric to assess performance.}}
        \renewcommand\arraystretch{1.05}
    \label{tab:para}
    \setlength{\tabcolsep}{2mm}{
    \begin{tabular}{c||ccccc}
        \toprule[1pt]
        p&0.00&0.25&0.50&0.75&1.00\\
        \cmidrule[0.3pt]{1-6}
        Fundus&87.12&88.64&88.72&\textbf{89.68}&89.16\\
        Prostate&87.82&88.93&89.64&\textbf{90.02}&88.37\\
        \midrule[1pt]
        \textcolor{black}{$\lambda$}&\textcolor{black}{0.05}&\textcolor{black}{0.10}&\textcolor{black}{0.20}&\textcolor{black}{0.50}&\textcolor{black}{1.00}\\
        \cmidrule[0.3pt]{1-6}
        \textcolor{black}{Fundus}&\textcolor{black}{89.20}&\textcolor{black}{89.68}&\textcolor{black}{\textbf{89.71}}&\textcolor{black}{89.35}&\textcolor{black}{89.04}\\
        \textcolor{black}{Prostate}&\textcolor{black}{89.79}&\textcolor{black}{\textbf{90.02}}&\textcolor{black}{89.67}&\textcolor{black}{89.06}&\textcolor{black}{88.96}\\

        \bottomrule[1pt]
    \end{tabular}
    }
\end{table}

\begin{table}
\centering
    \caption{\textcolor{black}{\textit{P}-value of paired \textit{t}-tests between Mamba-Sea and others for the \textbf{Fundus} and \textbf{Prostate} datasets on the sample level.}}
        \renewcommand\arraystretch{1.05}
    \label{tab:t-test}
    \setlength{\tabcolsep}{0.4mm}{
    {\scriptsize
    \textcolor{black}{\begin{tabular}{c||c|c|c|c|c}
        \toprule[1pt]
        Dataset & RAM-DSIR\cite{13} & DCAC\cite{hu2022domain} & WT-PSE\cite{chen2024learning} & SAM-Med2D\cite{cheng2023sam} & Med-SA\cite{wu2023medical} \\
        \cmidrule[0.3pt]{1-6}
        Fundus&8.05E-05&1.41E-08&8.71E-13&7.91E-10&4.12E-27\\
        Prostate&7.71E-09&1.66E-10&3.27E-08&9.94E-09&1.85E-08\\
        \cmidrule[1pt]{1-6}
        Dataset & Baseline & w/o GVA & w/o LSA & \multicolumn{2}{c}{w/o Consistent Training} \\
        \cmidrule[0.3pt]{1-6}
        Fundus&5.00E-14&8.32E-04&1.02E-06&\multicolumn{2}{c}{9.46E-04}\\
        Prostate&7.77E-12&3.56E-07&5.94E-09&\multicolumn{2}{c}{1.86E-06}\\
        \bottomrule[1pt]
    \end{tabular}
    }}}
\end{table}

\begin{table*}
\setlength{\abovecaptionskip}{0.2pt}
\centering
    \caption{{\textcolor{black}{Dice coefficient [\%], ASD [voxel] statistical analyse and params [M] of different methods on Skin Lesion Segmentation dataset. Top results are highlighted in \textbf{bold}.}}}
    \renewcommand\arraystretch{1.05}
    \label{tab:skin}
    \setlength{\tabcolsep}{2mm}{
        \begin{threeparttable}
        \textcolor{black}{
    \begin{tabular}{c|c||>{\centering\arraybackslash}p{1.5cm}|>{\centering\arraybackslash}p{1.5cm}|>{\centering\arraybackslash}p{1.5cm}|>{\centering\arraybackslash}p{1.5cm}}
        \toprule[1pt]
        \multicolumn{2}{c||}{Task}&\multicolumn{4}{c}{Skin Lesion Segmentation} \\
        \cmidrule[0.3pt]{1-6} 
        \multicolumn{2}{c||}{Model}&Dice $\uparrow$&ASD $\downarrow$&\textit{p}-value\tnote{*}&Params $\downarrow$  \\       
        \cmidrule[0.3pt]{1-6} 
        \multirow{3}{*}{Previous DG Methods}&\text{\tiny 21'}FedDG \cite{liu2021feddg}&92.06&4.91&1.13E-09&\textbf{3.18}\\
        &\text{\tiny 22'}RAM-DSIR \cite{13}&91.86&5.59&1.38E-06&3.80\\
        &\text{\tiny 24'}WT-PSE \cite{hu2022domain}&92.12&5.58&5.57E-05&9.57\\
        \cmidrule[0.3pt]{1-6} 
        \multirow{3}{*}{SAM-based Methods}&\text{\tiny 23'}SAM-Med2D \cite{chen2024learning}&90.60&6.45&2.47E-18&271.24\\
        &\text{\tiny 23'}Medical SAM Adapter \cite{cheng2023sam}&92.37&4.92&3.98E-07&636\\
        &\text{\tiny 24'}ESP-MedSAM \cite{xu2024esp}&91.45&6.03&1.70E-15&28.46\\
        \cmidrule[0.3pt]{1-6} 
        \multirow{2}{*}{Pure Backbone}&Swin-UNet \cite{2}&90.73&5.26&6.84E-08&27.17\\
        &VM-UNet \cite{ruan2024vm}&90.78&5.72&4.57E-16&27.43\\
        \cmidrule[0.3pt]{1-6} 
        \rowcolor{lightgray}
        Ours&\textbf{Mamba-Sea} &\textbf{93.11}&\textbf{3.88}&-&27.43\\
        \bottomrule[1pt]
    \end{tabular}}
    \begin{tablenotes}
\footnotesize
\item[\textcolor{black}{*}] \textcolor{black}{Paired T-tests Between Mamba-Sea And Others For Skin Lesion Segmentation On The Sample Level.}
\end{tablenotes}
    \end{threeparttable}}
\end{table*}

\begin{figure*}[tp]  
    \includegraphics[width=\linewidth]{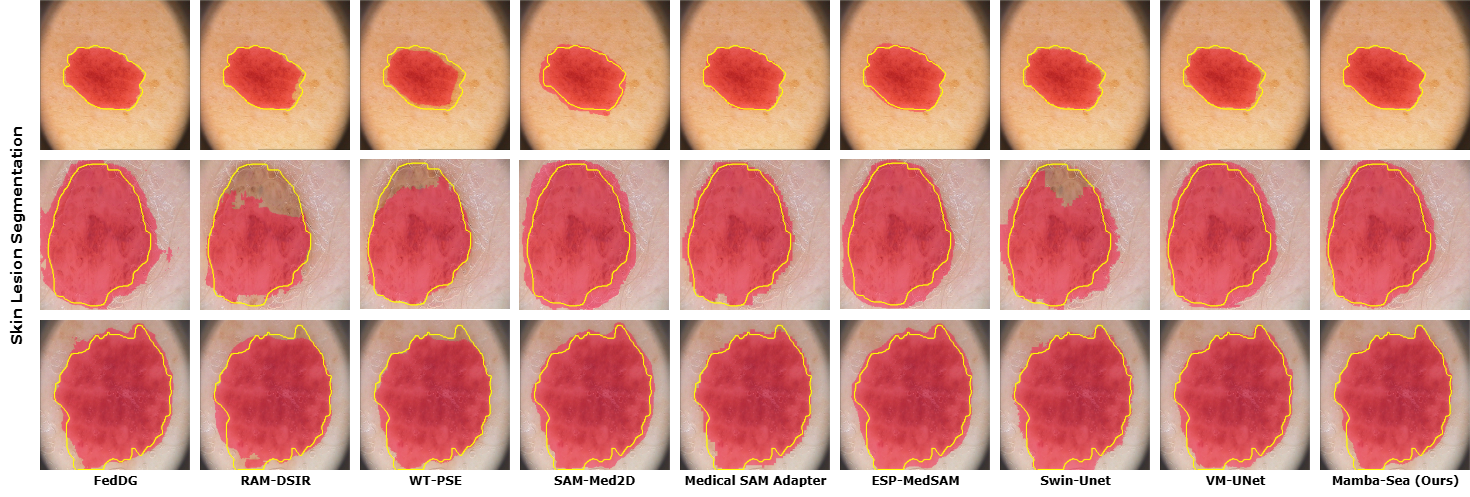}
    \caption{
    Visualization on segmentation results of different methods on \textbf{Skin} lesion segmentation dataset. The yellow contours indicate the boundaries of ground truths while the semi-transparent overlays are predictions.} 
    \label{mask3}
\end{figure*}

\subsubsection{Impact of Hyperparameters}
{As shown in Tab. \ref{tab:para}, we further investigate the impact of hyperparameters on model performance. 
Our Mamba-Sea employs sequence-wise style augmentation by applying style transformation to continuous regions defined by $p$ \textcolor{black}{in Eq. \eqref{mask2_local} and balances the weights of the segmentation loss and consistency loss by $\lambda$ in Eq. \eqref{eq:total_loss}.}  
We observe consistent performance improvements across different values of $p$, underscoring the effectiveness of sequence-wise style augmentation compared to the absence of this module. 
Extensive experiments reveal that the optimal value is $0.75$. At this value, we analyze that the model experiences sufficient distribution perturbation, which enhances generalization while minimizing the interference of excessive noise in the learning process. 
Additionally, we conduct experiments by varying the values of 
$\lambda$ and observe that these variations have minimal impact on the overall performance of our method. These findings demonstrate that our method is relatively stable with respect to changes in hyperparameter settings. To ensure consistency and fairness in our evaluations, we maintain $\lambda = 0.1$ across all datasets throughout our experiments. 

}

\subsubsection{Statistical Analysis}
\textcolor{black}{To assess the statistical significance of our method's performance improvements, following previous works\cite{14,chen2024learning}, we conduct paired \textit{t}-tests on both the Fundus and Prostate datasets. These tests are specifically designed to determine whether improvements achieved by our method are statistically superior to those of comparative approaches. The statistical analysis results, presented in Tab.~\ref{tab:t-test}, demonstrate that the observed enhancements in Dice scores are statistically significant at the conventional level of $p \textless 0.05$. Furthermore, to validate the individual contributions of our framework's components, we perform additional \textit{t}-tests comparing the performance outcomes of LSA and GVA modules, thereby providing quantitative evidence for the effectiveness of each architectural component.}

\subsubsection{Experiments on large-scale dataset}
{\textcolor{black}{Given the inherent advantages of Mamba's architecture, particularly its input-dependent parameters that enhance sequence modeling capabilities compared to traditional recurrent neural networks, coupled with its linear time complexity that ensures computational efficiency, we hypothesize that Mamba's potential is particularly motivated when processing large-scale datasets. This architectural superiority is expected to convert into enhanced generalization capabilities. 
To empirically validate this hypothesis, we extend our experimental evaluation to include a large-scale skin lesion segmentation dataset (S1 → T1) from ESP-MedSAM\cite{xu2024esp}, which is an efficient self-prompting SAM for DG in medical image segmentation. This comprehensive dataset comprises $3,694$ training images \cite{tschandl2018ham10000,codella2019skin} and $200$ testing images \cite{mendoncca2013ph}, providing a robust platform for assessing Mamba's performance in large-scale medical image segmentation tasks.
For comparative analysis, we select three SOTA DG methods, three representative SAM-based approaches and a ViT-based backbone architecture. As demonstrated in Tab. \ref{tab:skin}, Mamba-Sea achieves superior performance with a Dice coefficient of $93.11\%$, outperforming the baseline by $2.33\%$ ($93.11\% \textit{ vs. } 90.78\%$) and surpassing the second-best performing method Med-SA by $0.74\%$ ($93.11\% \textit{ vs. } 92.37\%$). All performance improvements are statistically significant ($p \textless 0.05$), demonstrating the effectiveness and generalizability of our proposed method in addressing domain shift challenges in medical image segmentation tasks. These results prove the universal applicability of our method's design in resolving domain shift problems prevalent in medical imaging. In Fig. \ref{mask3}, we visualize the segmentation results of our method alongside several approaches for medical images on the Skin lesion datasets.}}

\subsubsection{Comparison with SAM-based models}
{\textcolor{black}{Recent advancements in foundation models like SAM\cite{kirillov2023segment} and SAM2\cite{ravi2024sam}, have demonstrated generalization capabilities while they train on expensive datasets. This observation has prompted us to investigate their performance on specific out-of-distribution medical image segmentation tasks characterized by significant domain shifts between training and testing data distributions. Through extensive experimental evaluation across three distinct medical imaging domains - Fundus, Prostate, and Skin lesion segmentation (as detailed in Tab. \ref{tab:results_fundusdice}, \ref{tab:results_prodice} and \ref{tab:skin}), we observe that despite the substantial scale of training data utilized by these foundation models, their performance in handling distribution shifts reveals considerable room for improvement. 
Taking the large-scale skin lesion segmentation dataset as a representative case study, our Mamba-Sea framework demonstrates significant performance improvements over ESP-MedSAM\cite{xu2024esp}, the original method that introduced this dataset and desgined to fine-tune SAM to deal with DG tasks in medical image segmentation. 
Specifically, Mamba-Sea achieves a $1.66\%$ enhancement in Dice coefficient ($93.11\% \textit{ vs. } 91.45\%$). 
For computational costs, our model maintains competitive resource utilization with $27.43$ M parameters and $65.37$ GFLOPs for $1024 \times 1024$ input resolution, compared to ESP-MedSAM's $28.46$ M parameters and $55.88$ GFLOPs. 
This performance-complexity trade-off analysis suggests that while Mamba-Sea requires a little higher computational resources, it achieves superior segmentation accuracy with fewer parameters, demonstrating its effectiveness in balancing model performance and computational efficiency.
This finding suggests that while these models exhibit strong generalization capabilities in certain scenarios, their effectiveness in medical image segmentation tasks with domain shifts remains suboptimal, highlighting the need for specialized architectural adaptations\cite{chen2023sam,zhang2024improving}.}}

\section{Discussion}
\textcolor{black}{In this work, motivated by the demonstrated success of Mamba in supervised learning tasks, we investigate its generalization capabilities in the context of medical image segmentation. 
We experiment and find that the input-dependent matrices in Mamba architecture tend to accumulate and amplify domain-specific features during the training process, potentially compromising cross-domain generalization.} To address this issue, we propose Mamba-Sea, a novel Mamba-based framework that integrates global and local augmentation strategies. 
Extensive experimental evaluations demonstrate that our method achieves superior performance compared to SOTA methods across two widely-used public domain generalization benchmarks for medical image segmentation and an additional large-scale dataset, while maintaining the desirable property of linear computational complexity.

\textcolor{black}{Despite demonstrating superior performance in handling domain shifts in medical image segmentation tasks, our Mamba-Sea framework exhibits several limitations that warrant further investigation. Regarding the characteristics of the Mamba architecture itself, we conduct a comprehensive evaluation of its computational efficiency, focusing on training and inference times, as well as computational resource requirements quantified through Parameters (M) and FLOPs (G). To establish a robust comparative baseline, we perform extensive experiments on the Fundus dataset, encompassing CNN-based architectures (UNet \cite{ronneberger2015u} and DeepLabV3+ \cite{chen2018encoder}), ViT-based models (Swin-UNet \cite{2} and TransUNet \cite{chen2024transunet}), and our baseline Mamba-based VM-UNet \cite{ruan2024vm}. 
\begin{table}
\centering
    \caption{\textcolor{black}{Training and Inference time [s] with 
 computational costs (Params [M] and FLOPs [G]) on Fundus dataset. }}
        \renewcommand\arraystretch{1.05}
    \label{tab:trainingtime}
    \setlength{\tabcolsep}{0.8mm}{
    \begin{tabular}{c||cc|cc|c}
        \toprule[1pt]
        Model&Training&Inference&Params& FLOPs&Dice \\
        \cmidrule[0.3pt]{1-6}
        UNet \cite{ronneberger2015u}&270.50&3.44&\textbf{1.81}&\textbf{2.98}&79.82\\
        DeepLabV3+ \cite{chen2018encoder}&\textbf{262.16}&\textbf{3.37}&40.47&19.07&83.76\\
           \cmidrule[0.3pt]{1-6}
        Swin-UNet \cite{2}&280.94&3.44&27.17&6.10&80.89\\
        TransUNet \cite{chen2024transunet}&278.05&3.43&105.32&38.46&84.57\\
        \cmidrule[0.3pt]{1-6} 
        VM-UNet \cite{ruan2024vm}&268.09&3.56&27.43&4.09&\textbf{86.90}\\
        \bottomrule[1pt]
    \end{tabular}
    }
    \end{table}
We systematically evaluate these models on the Fundus dataset, documenting their average training and testing times across four distinct domains in Tab. \ref{tab:trainingtime}. All timing measurements are conducted under standardized conditions: training times are measured on a V$100$ GPU with a batch size of $8$, running a single task for $800$ iterations, while inference times are recorded on the same hardware with a batch size of $1$. 
The experimental results indicate that Mamba's training time remains competitive with CNN and ViT-based models, despite achieving significant performance improvements. 
\textcolor{black}{For example, although ViT-based Swin-Transformer architecture captures long-range interactions and exhibits linear computational complexity \cite{liu2021swin}, it still introduces significant computational costs because of its self-attention mechanism \cite{xing2024segmamba}. As shown in Tab. \ref{tab:skin} and Tab. \ref{tab:trainingtime}, we find that Swin-UNet \cite{2}, which is based on Swin-Transformer, still has relatively higher FLOPs and longer training time with a decline in performance compared with Mamba-Sea.} 
However, as evidenced Tab. \ref{tab:trainingtime} and \ref{tab:skin}, VM-UNet exhibits a relatively higher parameter count compared to some previous models, presenting an opportunity for further architectural optimization.}

Motivated by these findings, we propose several potential directions for enhancing the Mamba architecture in DG for medical image segmentation. To address computational efficiency, we plan to investigate the integration of advanced computing techniques, building upon recent advancements in Mamba2 \cite{mamba2} and Famba-V \cite{shen2024famba}. 
Furthermore, we aim to explore the incorporation of lightweight model architectures, potentially leading to the development of Ultralight VM-UNet\cite{wu2024ultralight}, which could significantly reduce computational overhead while maintaining competitive performance.
Lastly, in consideration of the incorporation of the inductive biases of locality and multiscale features in Swin-Transformer \cite{liu2021swin}, exploring how to combine the advantages of Swin-Transformer and Mamba to effectively address domain shifts in medical image segmentation is also meaningful.

In addition to limitations of the Mamba architecture itself, we have conducted a thorough analysis of the limitations specific to our proposed approach and identified potential directions for future enhancements. Specifically, we have integrated the structural characteristics of Mamba with global-to-local augmentations to enhance model generalization in medical image segmentation under domain shifts. 
We employ style perturbation to disrupt domain-specific information within salient tokens of input sequences. However, we recognize that various forms of domain-specific information may persist in real-world medical images, which our method may not completely mitigate. 
To address this limitation, we propose potential solutions, such as the development of advanced feature disentanglement methods that can adaptively distinguish and perturb domain-specific information. 
Furthermore, designing class-aware feature augmentation strategies could help reduce the accumulation of domain-related features. 

We would also like to explore the integration of our Mamba-based DG method with foundation models, recognizing their complementary strengths. Existing foundation models typically achieve promising performance under the assumption of similar training and testing distributions\cite{chen2023sam,zhang2024improving}, while in medical imaging scenarios the clinical data resources often have significant distribution shifts.  
However, compared to foundation models like MedSAM \cite{ma2024segment}, the scale of datasets we used is still limited. Our current implementation of Mamba-Sea primarily explores Mamba's generalization capabilities on relatively small-scale medical imaging datasets. In our future research, we plan to investigate two directions: (1) the potential of developing Mamba as a foundation model through training on large-scale datasets, and (2) further exploration of its generalization performance enhancement when applied to extensive clinical datasets. The challenge of enabling large foundation models to effectively learn domain-invariant representations while maintaining robust performance on unseen domains remains an open research question with potential for exploration and innovation.

\section{Conclusion}
In this paper, we explore the generalization ability of Mamba in medical image segmentation tasks. 
Our study has systematically addressed three critical aspects: (1) why select Mamba as a baseline architecture for DG tasks in medical image segmentation; (2) the existing limitations of Mamba in addressing DG challenges; (3) how to design Mamba-based modules for DG in medical image segmentation. 
To address these issues, we propose Mamba-Sea with global and local augmentation. Mamba-Sea incorporates global feature augmentation through GVA and local style augmentation through LSA. We further investigate a semantic training strategy to promote prediction consistency between outputs generated from GVA and original images. We perform comprehensive experiments on three datasets to demonstrate the improvements and effectiveness of our Mamba-Sea. Hope our work inspires further research in DG for medical image segmentation and contributes valuable insights to the community.

\bibliographystyle{IEEEtran}
\bibliography{reference}

\end{document}